\date{}
\documentclass{article}

\usepackage[linesnumbered,ruled]{algorithm2e}
\usepackage{amsmath} 
\usepackage{comment}

\usepackage[a4paper]{geometry}
\usepackage{apacite}
\usepackage{natbib}
\usepackage{float}
\usepackage{graphicx} 
\usepackage{latexsym,amsmath,amsfonts,amsthm,amssymb,graphicx,color,mathrsfs}
\usepackage{xcolor}
\usepackage{subfigure}
\usepackage{subcaption}
\usepackage{tabularx}
\usepackage{booktabs}
\usepackage{bm}
\usepackage{tikz}
\usepackage{multirow} 
\usepackage{adjustbox}
\usepackage{array} 
\usepackage{listings}%
\usepackage{amssymb}
\numberwithin{equation}{section}
\theoremstyle{plain}

\theoremstyle{plain}

\theoremstyle{definition}

\usepackage{caption}
\usepackage[nottoc]{tocbibind}
\allowdisplaybreaks[4]
\usepackage{dsfont}
\DeclareMathAlphabet{\mathcal}{OMS}{cmsy}{m}{n}

\providecommand{\lemmaname}{Lemma}

\providecommand{\theoremname}{Theorem}

\makeatother

\providecommand{\definitionname}{Definition}
\providecommand{\lemmaname}{Lemma}
\providecommand{\theoremname}{Theorem}

\usepackage{authblk}

\begin{document}
\title{Image Inpainting via Stochastic Dynamics}

\author[1]{Jiaqi Kuang\thanks{Corresponding author: \texttt{jiaqi.kuang@oscar.ox.ac.uk}}}
\author[2]{Zihao Guo\thanks{Email: \texttt{guozh@gbu.edu.cn}}}
\author[1,3]{Zhongmin Qian\thanks{Email: \texttt{zhongmin.qian@maths.ox.ac.uk}}}

\affil[1]{Oxford Suzhou Centre for Advanced Research, University of Oxford,
Suzhou, China}

\affil[2]{Institute for Advanced Research, Great Bay University,
Dongguan, Guangdong, China}

\affil[3]{Mathematical Institute, University of Oxford,
Andrew Wiles Building, Radcliffe Observatory Quarter,
Oxford, OX2 6GG, Oxfordshire, United Kingdom}

\date{}


\date{}

\maketitle
\begin{abstract}
Image inpainting aims to recover missing regions while preserving structural consistency. We propose a non-parametric method without network training based on data-guided stochastic dynamics. Starting from a masked image, the missing pixels are evolved through a reverse-time stochastic differential equation with a kernel-weighted correction estimated directly from a reference dataset. This empirical correction guides the reconstruction toward high-density regions of the data distribution without training a neural network or fitting a parametric density model. Experiments on MNIST, Fashion-MNIST, and MVTec show that the proposed method outperforms Mean Fill, Telea, and Navier-Stokes inpainting in PSNR, SSIM, and visual quality. On CelebA, it remains competitive and produces plausible completions for structure-sensitive occlusions. These results demonstrate the effectiveness of empirical reference statistics as a non-parametric prior for image inpainting.

\medskip{}
\end{abstract}

\newpage

\section{Introduction}
Image inpainting—the task of reconstructing missing or corrupted regions in an image—has long served as a fundamental problem in computer vision and image processing. With the advancement of digital imaging technologies, the demand for reliable restoration has grown across diverse domains, including object removal in photo editing \citep{elharrouss2020image}, occlusion recovery in medical diagnostics \citep{li2016image}, defect correction in satellite imagery \citep{dong2018inpainting}, and even digital forensics \citep{zeng2017image}. The core challenge lies not merely in filling the missing region with visually plausible values, but in ensuring that the reconstructed content is consistent with both local image structures (e.g., edges, contours) and global statistical regularities inherent to the underlying data distribution—a requirement that becomes increasingly difficult as the size or semantic complexity of the missing region grows \citep{sridevi2019image, li2017localization}.

Early approaches can be broadly categorized into two paradigms: diffusion-based and exemplar-based methods. Diffusion-based techniques, pioneered by Bertalmio et al. \citep{bertalmio2000image}, propagate information from known boundaries into the missing region using partial differential equations (PDEs), effectively “smearing” isophotes across the hole. While effective for small, smooth regions, these methods suffer from blurriness and fail to recover complex textures or repetitive patterns \citep{li2016image}. To address this limitation, exemplar-based methods such as Criminisi’s algorithm \citep{antonio2004region} were introduced, which reconstruct missing content by copying and pasting similar patches from the known image area. This strategy excels at preserving texture but often struggles with semantic coherence in highly structured scenes (e.g., faces, text, or man-made objects), where global layout constraints must be respected \citep{elharrouss2020image}. Subsequent improvements combined both paradigms—e.g., Telea’s fast marching method \citep{telea2004image} blends diffusion with confidence-driven propagation—yet remain fundamentally local and heuristic.

The advent of deep learning has revolutionized the field. Convolutional Neural Networks (CNNs) and Generative Adversarial Networks (GANs) now dominate state-of-the-art performance, capable of generating semantically meaningful and visually realistic content \citep{elharrouss2020image}. Context encoders \citep{pathak2016context} first demonstrated end-to-end trainable inpainting, while GAN-based models like DeepFill \citep{yu2018generative} introduced contextual attention to capture long-range dependencies. More recently, diffusion-based generative models grounded in stochastic differential equations have achieved remarkable fidelity by modeling the data distribution through iterative denoising \citep{song2020score}. However, these machine learning approaches typically require extensive training on large-scale datasets, are computationally intensive, and may overfit to specific object categories or scene types. Moreover, their black-box nature may limit interpretability, and the generated content may not always remain consistent with the observed image structure.

From a probabilistic perspective, image inpainting is naturally a
conditional generation problem rather than a purely deterministic
interpolation task. Given the observed pixels, the objective is to
recover missing content that is not only locally compatible with the
visible boundary, but also likely under the underlying image
distribution. Forward-reverse diffusion provides a natural mechanism
for this purpose. A forward diffusion progressively smooths a complex
data distribution through stochastic perturbations, whereas the
corresponding reverse-time dynamics transports noisy or incomplete
states toward regions of high data probability. For image inpainting,
the observed pixels can be enforced as hard constraints throughout the
reverse evolution, while only the missing coordinates are updated.
This makes reverse diffusion particularly suitable for combining
observation consistency with distribution-level structural priors.
Existing diffusion-based inpainting methods typically learn the
reverse-time score using a neural network. In contrast, we investigate
whether the required distributional guidance can be estimated directly
from a finite reference set, thereby retaining the stochastic
forward--reverse formulation without training a parametric score model.

The forward--reverse stochastic viewpoint adopted here is also closely
related to a broader program of probabilistic representations for
partial differential equations developed by Qian and collaborators.
This probabilistic framework was subsequently applied to fluid
dynamical equations. In particular, Cruzeiro and Qian
\citep{cruzeiro2014backward} derived a nonlinear Feynman--Kac
representation for the two-dimensional vorticity equation and formulated
the corresponding fluid evolution through a stochastic terminal-value
problem. More recently, Qian \citep{qian20222} developed
stochastic representations for incompressible Navier-Stokes flows in
wall-bounded domains, providing a mathematical basis for Monte Carlo
simulation of viscous and turbulent flows. Related stochastic particle
formulations have also been used to construct random-vortex-type
computational methods for more general fluid systems
\citep{Cherepanov2024Boundary,Guo2025}. These works demonstrate that forward and
backward diffusion processes are not merely formal probabilistic tools,
but can serve as practical mechanisms for reconstructing solutions of
high-dimensional evolution equations from stochastic trajectories.
Motivated by this perspective, we adapt the time-reversal diffusion
framework to image inpainting, where the distribution-dependent drift
is approximated directly from an empirical measure supported on a
reference image dataset.

Despite these advances, there remains significant interest in non-parametric, training-free inpainting strategies—especially in low-data regimes, privacy-sensitive applications, or scenarios demanding algorithmic transparency. Recent works have revisited classical frameworks with modern twists: Li et al. \citep{li2016image} enhanced the Total Variation (TV) model with evolutionary optimization for patch selection; Sridevi and Kumar \citep{sridevi2019image} proposed a fractional-order nonlinear diffusion model to mitigate staircase artifacts; and Zhang et al. \citep{zhang2020deeply} integrated sparse representation with geometric priors for remote sensing images. Notably, non-learning methods that leverage dataset-level statistics without explicit modeling remain underexplored. Most existing non-learning approaches—including kernel-based interpolation \citep{takeda2007kernel}, low-rank modeling \citep{liu2012exemplar, ruvzic2014context, jin2015annihilating}, steering-kernel-based patch matching \citep{ghorai2019multiple}, and total variation models \citep{rudin1992nonlinear,li2016image}—rely fundamentally on \textit{within-image redundancy}: they assume the missing content can be inferred from the observed part of the same image through self-similarity, smoothness, or global structural priors. However, this assumption often fails in scenarios involving small, structured objects (e.g., handwritten digits or fashion items) where intra-image repetition is limited. In contrast, when a reference dataset of similar instances is available—such as a collection of digit images or clothing samples—it becomes possible to exploit \textit{cross-instance statistical regularities} without any model training. Yet, non-parametric, training-free inpainting strategies that explicitly harness such external dataset-level priors remain largely unexplored.

In this work, we propose a non-parametric image inpainting method based on data-guided reverse stochastic dynamics. Starting from a masked observation, the missing coordinates are evolved through a sequence of stochastic updates, while the observed pixels are clamped to their original values throughout the reconstruction. At each iteration, the distribution-dependent drift is approximated by a kernel-weighted average of deviations between the current state and samples from a reference dataset, thereby guiding the trajectory toward high-density regions of the empirical data distribution without training a neural network or explicitly fitting a parametric density model.

The main contributions of this work are threefold. First, we formulate image inpainting as a conditional stochastic reconstruction problem within a forward--reverse diffusion framework. Second, we construct a non-parametric empirical approximation of the reverse drift directly from reference samples, without neural-network training, and introduce a smooth regularization of the singular Ornstein--Uhlenbeck bridge drift for stable numerical computation. Third, we evaluate the proposed method on MNIST
\citep{lecun1998gradient}, Fashion-MNIST
\citep{xiao2017fashion}, MVTec AD, and CelebA. The results show that the method produces structurally coherent reconstructions and outperforms Mean Fill and classical PDE-based approaches, including Telea \citep{telea2004image} and Navier--Stokes inpainting, under the considered experimental settings.

The structure of this paper is organized as follows: Section~\ref{theory} presents the theoretical formulation. Section~\ref{method} describes the inpainting method and experimental setup. Section~\ref{results} reports results on MNIST, Fashion-MNIST, MVTec AD, and CelebA, together with ablation studies. Section~\ref{conclusion} concludes the paper.

\section{Theory Foundation}
\label{theory}

The backward stochastic framework used here is closely related to the general theory of backward stochastic dynamics developed by Qian and his collaborators ~\citep{liang2011backward,qian20222}, which is widely used in fluid dynamic \citep{qian2022,Guo2025}. Let $\mu$ be an unknown probability distribution on $\mathbb{R}^{D}$, from which we are given an i.i.d.\ sample $\{\xi_{n}\}_{n=1}^{N}$. Our goal is to construct a reverse-time stochastic dynamics that starts from a prescribed endpoint $\eta\in\mathbb{R}^{D}$ and terminates at a random state whose law coincides with $\mu$, without explicitly modeling the density of $\mu$. To this end, we consider a forward Ornstein--Uhlenbeck (OU) process $(Z_{t})_{t\in[0,T]}$ defined by
\begin{equation}
    dZ_{t}=\sqrt{2\nu}\,dB_{t}+\beta Z_{t}\,dt,\quad Z_{0}\sim\mu,
    \label{eq:forward_ou}
\end{equation}
where $B_{t}$ is a standard $D$-dimensional Brownian motion, $\nu>0$ is the diffusion coefficient, and $\beta\in\mathbb{R}$ is the drift parameter. The infinitesimal generator of~\eqref{eq:forward_ou} is $L=\nu\Delta+\beta x\cdot\nabla$, and the corresponding transition density is Gaussian:
\begin{equation}
p(s,x;t,y)
=
\Biggl(2\pi\frac{\nu}{\beta}\bigl(e^{2\beta(t-s)}-1\bigr)\Biggr)^{-D/2}
\exp\!\Biggl(
-\frac{\beta}{2\nu}
\frac{\|y-e^{\beta(t-s)}x\|^{2}}{e^{2\beta(t-s)}-1}
\Biggr),
\qquad 0\le s<t.
\label{eq:ou_transition}
\end{equation}

Conditioning the forward process on the terminal value $Z_{T}=\eta$ yields a conditional diffusion bridge. Time reversal of this bridge produces a reverse-time process $(Y_{t})_{t\in[0,T]}$ with $Y_{0}=\eta$, whose drift consists of the time-reversed OU contribution together with a score term that reintroduces the unknown measure $\mu$ through the transition-density ratio. The resulting dynamics take the form
\begin{equation}
    dY_{t}=\sqrt{2\nu}\,dB_{t}-\beta Y_{t}\,dt+2\nu\nabla_{y}\log\left(\int_{\mathbb{R}^{D}}\frac{p(t,\xi;T,y)}{p(0,\xi;T,\eta)}\,\mu(d\xi)\right)\bigg|_{y=Y_{t}}dt,
    \label{eq:reverse_general}
\end{equation}
where $p(s,x;t,y)$ is given by~\eqref{eq:ou_transition}. The integral against $\mu$ encodes the influence of the data distribution and is retained as a formal object when $\mu$ is unknown.

Because $p$ is Gaussian, the score in~\eqref{eq:reverse_general} admits an explicit kernel representation. Substituting~\eqref{eq:ou_transition} and completing the square shows that
\begin{equation}
\frac{p(t,\xi;T,y)}{p(0,\xi;T,\eta)}
\propto
\exp\!\left(-\frac{\|\xi-M(y,t)\|^{2}}{2\rho(t)^{2}}\right),
\label{eq:kernel_form}
\end{equation}
up to a factor independent of $\xi$, where
\begin{align}
    M(y,t)&=e^{-\beta(T-t)}y+\frac{e^{\beta(T-t)}-e^{-\beta(T-t)}}{e^{\beta t}-e^{-\beta t}}\left(e^{\beta t}y-\eta\right),
    \label{eq:M}
    \\
    \rho(t)^{2}&=\frac{\nu}{\beta}\cdot\frac{e^{2\beta(T-t)}-1}{e^{2\beta(T-t)}}\cdot\frac{e^{2\beta T}-1}{e^{2\beta t}-1}.
    \label{eq:rho}
\end{align}
Here $M(y,t)$ is the conditional mean associated with the OU transition, and $\rho(t)$ is the corresponding scale. Inserting~\eqref{eq:kernel_form} into~\eqref{eq:reverse_general} and collecting the linear terms yields the closed-form reverse SDE
\begin{equation}
\begin{aligned}
dY_{t}=&\sqrt{2\nu}\,dB_{t} \\
&+\frac{\beta}{e^{2\beta t}-1}\left((e^{2\beta t}+1)Y_{t}-2e^{\beta t}\eta\right)dt \\
&+\frac{2\beta e^{\beta(T-t)}}{e^{2\beta(T-t)}-1}\cdot\frac{
    \displaystyle\int_{\mathbb{R}^{D}}(\xi-M(Y_{t},t))\exp\left(-\frac{\|\xi-M(Y_{t},t)\|^{2}}{2\rho(t)^{2}}\right)\mu(d\xi)
}{
    \displaystyle\int_{\mathbb{R}^{D}}\exp\left(-\frac{\|\xi-M(Y_{t},t)\|^{2}}{2\rho(t)^{2}}\right)\mu(d\xi)
}\,dt,
\end{aligned}
\label{eq:sde_final_theory}
\end{equation}
The first drift term is determined by the OU prior conditioned on the endpoint $\eta$, while the second drift term provides a kernel-weighted correction driven by $\mu$. Together, these two contributions yield the reverse-time dynamics used in our method. 

Equation~\eqref{eq:sde_final_theory} is the theoretical foundation of our method. In practice, $\mu$ is replaced by the empirical measure induced by $\{\xi_{n}\}_{n=1}^{N}$, so that the integrals become finite sums over reference samples. For image inpainting, we apply the dynamics in~\eqref{eq:sde_final_theory} by using the masked observation as the initial state and the empirical mean of the retrieved reference pool as the endpoint parameter. The numerical implementation is detailed in Section~\ref{method}.

\section{Method}
\label{method}
\subsection{Model}
Based on Section~\ref{theory}, we propose a non-parametric image inpainting method based on a reverse-time SDE guided by the empirical distribution of a reference dataset. No neural network is trained. Starting from a masked observation, the reverse dynamics update the missing pixels using an empirical correction computed from reference samples.

Let $y_{\mathrm{obs}}$ denote the masked observation, with missing entries filled by a placeholder value. Given a retrieved reference pool $\{\xi_n\}_{n=1}^{N_r}$, we define its empirical mean as
\begin{equation}
\eta_{\mathrm{ref}}
=
\frac{1}{N_r}\sum_{n=1}^{N_r}\xi_n.
\label{eq:eta_ref}
\end{equation}
The practical dynamics are initialized by
\begin{equation}
Y_0=y_{\mathrm{obs}}.
\label{eq:method_initial}
\end{equation}

At each SDE step, a finite subset $\{x^{(k)}\}_{k=1}^{K}$ is sampled from the retrieved reference pool $\{\xi_n\}_{n=1}^{N_r}$.

Specifically, the following SDE is used for the latent variable $Y_t\in\mathbb{R}^d$:

\begin{equation}
\begin{aligned}
dY_t
&=
\sqrt{2\nu}\,dB_t
+
\underbrace{
\frac{\beta}{e^{2\beta t}-1}
\left[
\left(e^{2\beta t}+1\right)Y_t
-
2e^{\beta t}\eta_{\mathrm{ref}}
\right]
}_{\text{linear drift}}dt
\\
&\quad+
\underbrace{
\frac{2\beta e^{\beta(T-t)}}{e^{2\beta(T-t)}-1}
\frac{
\bar U_1\left((x^{(k)})_{k=1}^{K};Y_t,t\right)
}{
\bar U_2\left((x^{(k)})_{k=1}^{K};Y_t,t\right)
}
}_{\text{data-driven correction}}dt.
\end{aligned}
\label{eq:method_sde}
\end{equation}

For the practical scheme, we define
\begin{equation}
M_{\mathrm{ref}}(Y_t,t)
:=
M(Y_t,t)\big|_{\eta=\eta_{\mathrm{ref}}},
\end{equation}
where $M$ is the OU conditional mean derived in Section~\ref{theory}.

The empirical functions $\bar U_1$ and $\bar U_2$ are then defined as

\begin{align}
\bar U_1\left((x^{(k)})_{k=1}^{K};Y_t,t\right)
&=
\sum_{k=1}^{K}
\exp\left(
-\frac{
\left\|x^{(k)}-M_{\mathrm{ref}}(Y_t,t)\right\|^2
}{
2\rho(t)^2
}
\right)
\left(
x^{(k)}-M_{\mathrm{ref}}(Y_t,t)
\right),
\label{eq:empirical_u1}
\\
\bar U_2\left((x^{(k)})_{k=1}^{K};Y_t,t\right)
&=
\sum_{k=1}^{K}
\exp\left(
-\frac{
\left\|x^{(k)}-M_{\mathrm{ref}}(Y_t,t)\right\|^2
}{
2\rho(t)^2
}
\right),
\end{align}

where $\rho(t)$ is the corresponding OU scale and
$\{x^{(k)}\}_{k=1}^{K}$ is a finite subset sampled from the retrieved reference pool.

The unregularized linear component in~\eqref{eq:method_sde} is obtained from the exact OU-bridge drift by setting its endpoint parameter to $\eta_{\mathrm{ref}}$. To identify the term responsible for the singular behavior near $t=0$, we rewrite this component as

\begin{equation}
\begin{aligned}
&\frac{\beta}{e^{2\beta t}-1}
\left[
\left(e^{2\beta t}+1\right)Y_t
-
2e^{\beta t}\eta_{\mathrm{ref}}
\right]
\\
&\qquad=
\beta\coth(\beta t)
\left(Y_t-\eta_{\mathrm{ref}}\right)
+
\beta\tanh\left(\frac{\beta t}{2}\right)
\eta_{\mathrm{ref}},
\end{aligned}
\label{eq:ou_drift_decomposition}
\end{equation}
where we have used the identity
\[
\coth(x)-\operatorname{csch}(x)
=
\tanh\left(\frac{x}{2}\right).
\]

This decomposition isolates the potentially singular component of the linear drift. As $t\to0$,
\[
\coth(\beta t)=\frac{1}{\beta t}+O(t),
\qquad
\tanh\left(\frac{\beta t}{2}\right)
=
\frac{\beta t}{2}+O(t^3).
\]
Thus, the potentially singular coefficient appears only in the term multiplying $Y_t-\eta_{\mathrm{ref}}$, whereas the second term remains bounded and vanishes as $t\to0$. Since the practical dynamics are initialized from $y_{\mathrm{obs}}$, which generally differs from $\eta_{\mathrm{ref}}$, direct evaluation in an explicit Euler-Maruyama scheme may cause large updates near $t=0$.

To stabilize the numerical integration, we introduce
\begin{equation}
t_{\varepsilon}
=
\sqrt{t^2+\varepsilon^2},
\qquad
\varepsilon>0,
\label{eq:regularized_time}
\end{equation}
and evaluate the singular $\coth$ factor at $t_{\varepsilon}$. The resulting regularized linear drift is
\begin{equation}
\beta\coth(\beta t_{\varepsilon})
\left(Y_t-\eta_{\mathrm{ref}}\right)
+
\beta\tanh\left(\frac{\beta t}{2}\right)
\eta_{\mathrm{ref}}.
\label{eq:regularized_linear_drift}
\end{equation}
We use $\varepsilon=10^{-3}$ in all experiments. The nonsingular $\tanh(\beta t/2)$ term remains evaluated at the original time $t$.

Since $t_{\varepsilon}\geq\varepsilon$, the regularized $\coth$ coefficient remains bounded over the computational interval. Furthermore, for every fixed $t>0$, $t_{\varepsilon}\to t$ as $\varepsilon\to0$. Thus, the regularized expression approaches the original OU-bridge drift away from the singular endpoint.

Combining the regularized linear drift with the empirical correction, the SDE implemented in our experiments is
\begin{equation}
\begin{aligned}
dY_t
&=
\sqrt{2\nu}\,dB_t
\\
&\quad+
\left[
\beta\coth(\beta t_{\varepsilon})
\left(Y_t-\eta_{\mathrm{ref}}\right)
+
\beta\tanh\left(\frac{\beta t}{2}\right)
\eta_{\mathrm{ref}}
\right]dt
\\
&\quad+
\frac{2\beta e^{\beta(T-t)}}{e^{2\beta(T-t)}-1}
\frac{
\bar U_1\left((x^{(k)})_{k=1}^{K};Y_t,t\right)
}{
\bar U_2\left((x^{(k)})_{k=1}^{K};Y_t,t\right)
}
\,dt.
\end{aligned}
\label{eq:implemented_sde}
\end{equation}

\subsection{Implementation and Evaluation Setup}

We discretize~\eqref{eq:implemented_sde} using the Euler--Maruyama scheme. At each step, we evaluate $M_{\mathrm{ref}}$ and $\rho$, compute the regularized linear drift~\eqref{eq:regularized_linear_drift}, estimate the data-driven correction, and update the missing coordinates.

The empirical correction term is approximated by Monte Carlo sampling from a finite reference pool. 

For datasets with a large number of reference samples, we first retrieve a candidate pool using local-context distance and then sample a subset at each SDE step. The local context is defined as the observed ring around the missing region, obtained by dilating the binary mask, and distances for sample retrieval and kernel weighting are computed on this context region rather than on the entire image. This strategy improves conditional matching while keeping the computation tractable.

Observed pixels are clamped to their original values throughout the SDE iterations, so the update is only applied to the missing region. Unless otherwise specified, images are normalized to $[-1,1]$ before the SDE simulation and denormalized to $[0,1]$ for evaluation. We use $T=1.0$ and $\beta=2.0$ in all experiments. Dataset-specific values of the diffusion coefficient $\nu$, candidate-pool size, subset size, and mask size are reported with the corresponding experimental settings.

We evaluate the proposed method on MNIST, Fashion-MNIST, MVTec, and CelebA. Since our method is training-free and relies only on reference samples, we focus on non-learning baselines that share the same inference-time setting. Specifically, we compare against Mean Fill, Telea's fast marching inpainting method, and the Navier-Stokes inpainting method implemented in OpenCV. No neural network training is performed for any method in our comparison.

\section{Results}
\label{results}

\subsection{Experiments on MNIST and Fashion-MNIST}

For MNIST and Fashion-MNIST, we use $28\times28$ grayscale images with a randomly placed $12\times12$ square mask. Following the implementation details above, we set the context kernel size to 3, the candidate pool size to 10000, the subset size to 1024, and $\nu=0.05$.

\subsubsection{Mixed-label Results}

We first evaluate the proposed training-free SDE-based inpainting method on MNIST and Fashion-MNIST. Since our method requires no neural network training and relies only on reference samples, we compare against non-learning baselines under the same inference-time setting: Mean Fill, Telea's fast marching method~\citep{telea2004image}, and the Navier-Stokes inpainting method implemented in OpenCV. Mean Fill represents the simplest statistical prior, while Telea and Navier-Stokes represent classical PDE-based local propagation. It is worth emphasizing that our method can run on CPU without GPU resources.

Tables~\ref{tab:mnist_all_100} and~\ref{tab:mnist_all_500} report the quantitative results on MNIST. On 100 randomly sampled test images, our method achieves a PSNR of $19.46$~dB and an SSIM of $0.8582$, outperforming Mean Fill ($16.05$~dB / $0.7030$), Telea ($17.47$~dB / $0.7895$), and Navier-Stokes ($17.72$~dB / $0.7868$). On 500 test samples, the method remains clearly better than all baselines, achieving $18.83$~dB PSNR and $0.8429$ SSIM. The slight decrease from 100 to 500 samples is expected because more challenging cases are included, while the consistently higher means confirm the robustness of the proposed approach.

\begin{table}[H]
\centering
\caption{Comparison of Image Inpainting Methods on MNIST (100 test samples)}
\label{tab:mnist_all_100}
\setlength{\tabcolsep}{18pt}
\begin{tabular}{l|c|c}
\hline
Method & PSNR (dB) & SSIM \\
\hline
SDE            & $19.46 \pm 7.26$ & $0.8582 \pm 0.0994$ \\
Mean Fill      & $16.05 \pm 3.18$ & $0.7030 \pm 0.0678$ \\
Telea          & $17.47 \pm 3.43$ & $0.7895 \pm 0.0726$ \\
Navier-Stokes  & $17.72 \pm 3.53$ & $0.7868 \pm 0.0731$ \\
\hline
\end{tabular}
\end{table}

\begin{table}[H]
\centering
\caption{Comparison of Image Inpainting Methods on MNIST (500 test samples)}
\label{tab:mnist_all_500}
\setlength{\tabcolsep}{18pt}
\begin{tabular}{l|c|c}
\hline
Method & PSNR (dB) & SSIM \\
\hline
SDE            & $18.83 \pm 6.93$ & $0.8429 \pm 0.0979$ \\
Mean Fill      & $15.76 \pm 3.21$ & $0.6910 \pm 0.0757$ \\
Telea          & $17.15 \pm 3.28$ & $0.7825 \pm 0.0730$ \\
Navier-Stokes  & $17.36 \pm 3.37$ & $0.7771 \pm 0.0743$ \\
\hline
\end{tabular}
\end{table}

\begin{figure}[H]
    \centering
    \includegraphics[width=1\textwidth]{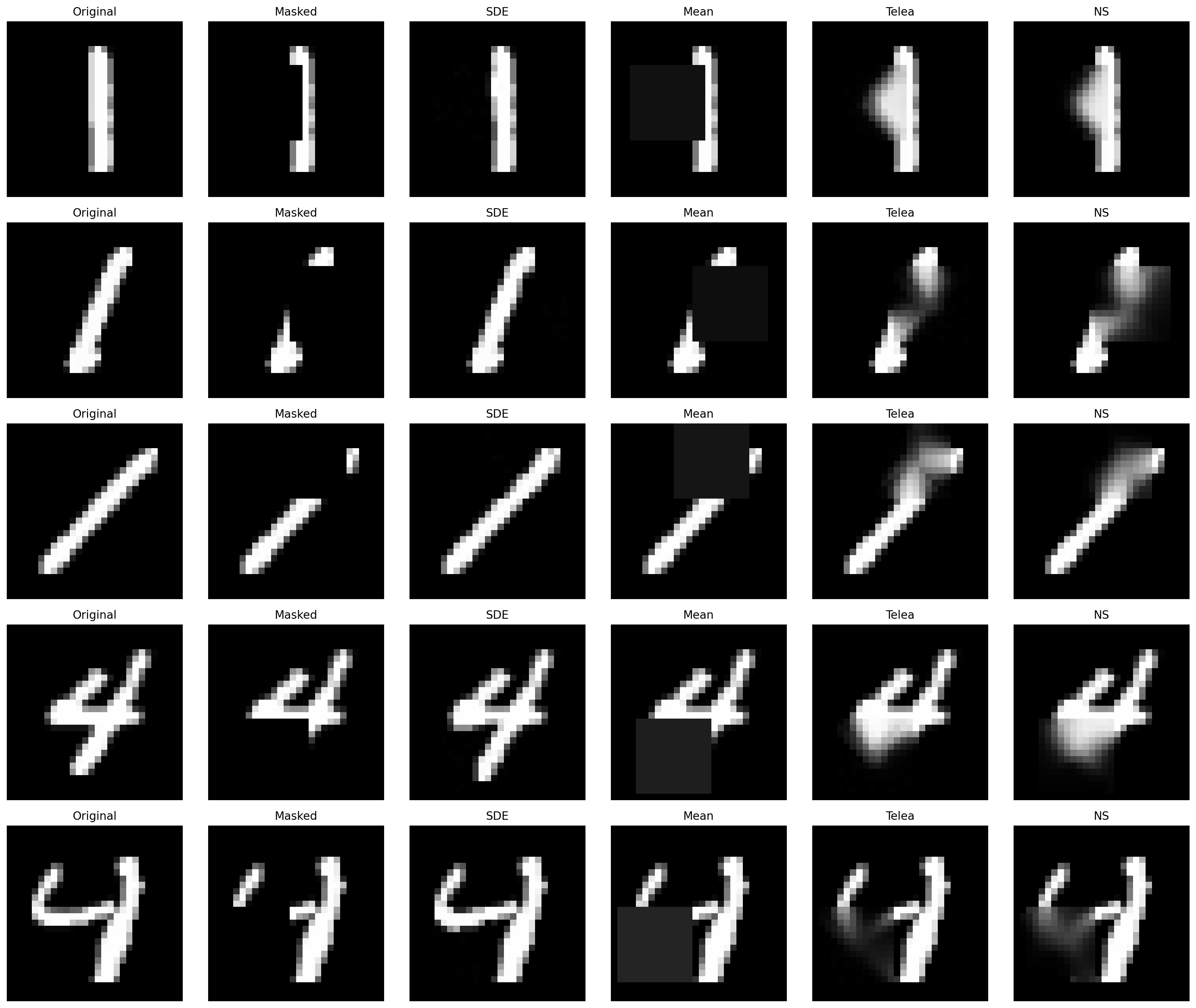}
    \caption{Mixed Label Comparison of Image Inpainting Performance of SDE, Mean-Fill, Telea, and NS in MNIST}
    \label{mnist_all}
\end{figure}

Figure~\ref{mnist_all} provides qualitative support for these results. Each row corresponds to a test instance with a randomly placed $12\times12$ square mask. Mean Fill consistently fails to recover meaningful content and leaves a flat patch in the missing region. Telea and Navier-Stokes produce smoother transitions, but often introduce blurring and geometric distortion, especially when the mask interrupts thin strokes or sharp corners. In contrast, the proposed SDE method reconstructs sharper stroke continuity and more coherent digit structures by leveraging an empirical prior from the reference dataset. For example, the vertical and slanted strokes of digit "1" and the corner structures of digit "4" are better preserved than by local PDE-based methods.

We further evaluate the method on Fashion-MNIST, which contains more complex object structures than handwritten digits. Tables~\ref{tab:fashion_mnist_all_100} and~\ref{tab:fashion_mnist_all_500} show that the proposed method again outperforms all baselines. On 100 test samples, it achieves a PSNR of $23.61$~dB and an SSIM of $0.8726$, compared with Mean Fill ($16.87$~dB / $0.7144$), Telea ($20.50$~dB / $0.8288$), and Navier-Stokes ($20.65$~dB / $0.8297$). On 500 samples, the performance remains stable at $23.63$~dB PSNR and $0.8739$ SSIM. The higher scores on Fashion-MNIST relative to MNIST suggest that the empirical-prior formulation is particularly effective for recovering structured object layouts under random occlusion.

\begin{table}[H]
\centering
\caption{Comparison of Image Inpainting Methods on Fashion-MNIST (100 test samples)}
\label{tab:fashion_mnist_all_100}
\setlength{\tabcolsep}{18pt}
\begin{tabular}{l|c|c}
\hline
Method & PSNR (dB) & SSIM \\
\hline
SDE            & $23.61 \pm 5.20$ & $0.8726 \pm 0.0773$ \\
Mean Fill      & $16.87 \pm 3.22$ & $0.7144 \pm 0.0647$ \\
Telea          & $20.50 \pm 3.37$ & $0.8288 \pm 0.0595$ \\
Navier-Stokes  & $20.65 \pm 3.40$ & $0.8297 \pm 0.0553$ \\
\hline
\end{tabular}
\end{table}

\begin{table}[H]
\centering
\caption{Comparison of Image Inpainting Methods on Fashion-MNIST (500 test samples)}
\label{tab:fashion_mnist_all_500}
\setlength{\tabcolsep}{18pt}
\begin{tabular}{l|c|c}
\hline
Method & PSNR (dB) & SSIM \\
\hline
SDE            & $23.63 \pm 5.43$ & $0.8739 \pm 0.0741$ \\
Mean Fill      & $16.79 \pm 3.27$ & $0.7029 \pm 0.0642$ \\
Telea          & $20.76 \pm 4.12$ & $0.8282 \pm 0.0559$ \\
Navier-Stokes  & $20.98 \pm 4.34$ & $0.8287 \pm 0.0559$ \\
\hline
\end{tabular}
\end{table}

\begin{figure}[H]
    \centering
    \includegraphics[width=1\textwidth]{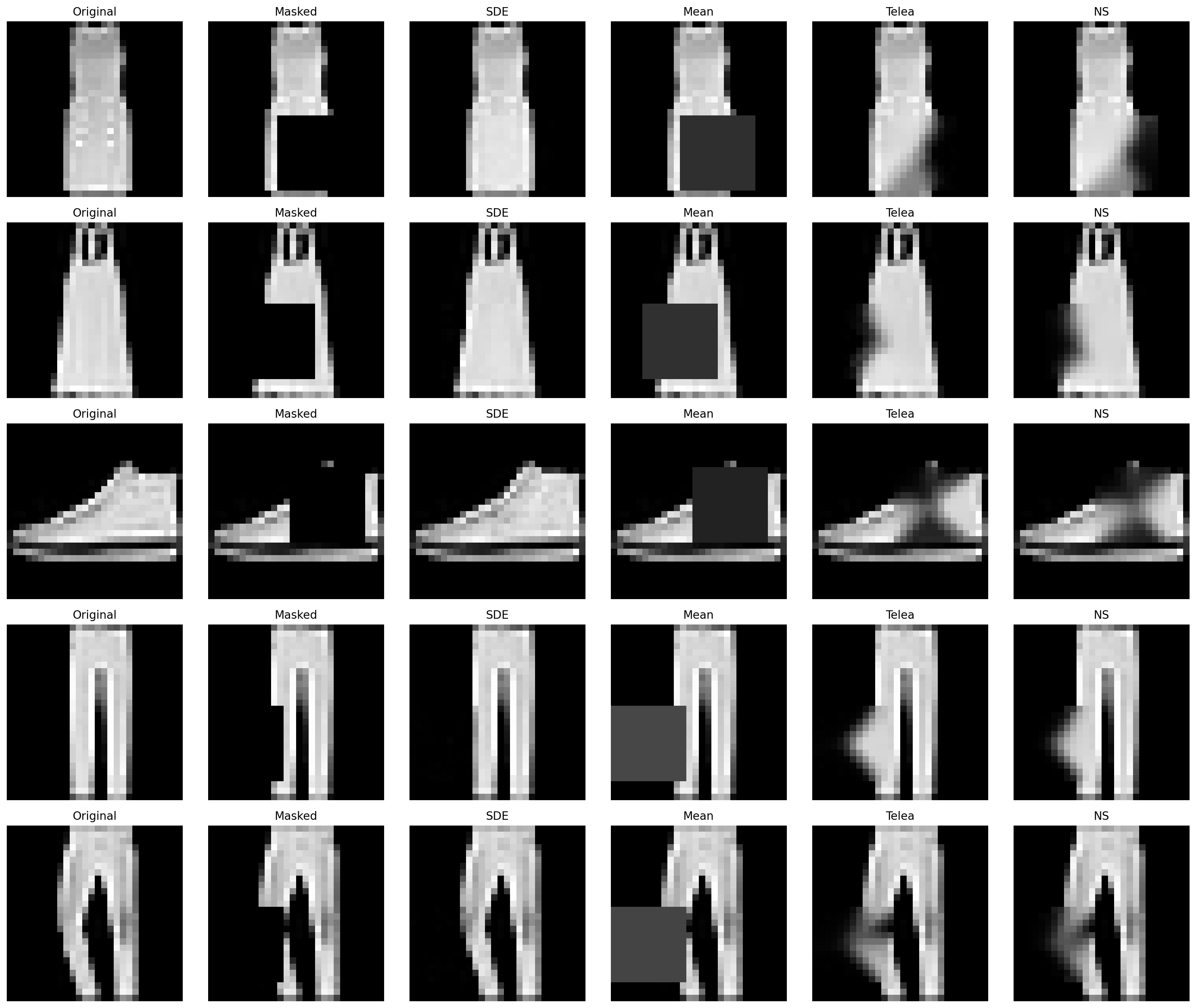}
    \caption{Mixed Label Comparison of Image Inpainting Performance of SDE, Mean-Fill, Telea, and NS in Fashion-MNIST}
    \label{fashionmnist_all}
\end{figure}

These quantitative gains are consistent with the qualitative results in Figure~\ref{fashionmnist_all}. Mean Fill fails to recover semantic content, while Telea and Navier-Stokes often blur object boundaries and distort geometry. By contrast, the proposed SDE method reconstructs sharper silhouettes and more plausible global shapes for clothing items such as dresses, sneakers, and trousers, even when a substantial portion of the object is occluded.

\subsubsection{Class-specific Results}

To further investigate the performance of our method under class-specific empirical priors, we conduct experiments on single-label subsets of MNIST and Fashion-MNIST, where the reference pool is restricted to the same class as the test samples

For MNIST, we focus on digits "1", "4", and "8", which represent thin strokes, diagonal/cross structures, and closed-loop topology, respectively. Tables~\ref{tab:mnist_class_1},~\ref{tab:mnist_class_4}, and~\ref{tab:mnist_class_8} report the quantitative results over 100 test samples for each class. The proposed SDE method consistently outperforms Mean Fill, Telea, and Navier-Stokes. The advantage is especially clear on digit "1", where the method achieves a PSNR of $28.25$~dB and an SSIM of $0.9345$, substantially higher than the local baselines. On digits "4" and "8", the method also remains the best, indicating that class-specific priors help recover both local stroke continuity and more complex digit topology.

\begin{table}[H]
\centering
\caption{Comparison on MNIST digit "1" (100 test samples)}
\label{tab:mnist_class_1}
\setlength{\tabcolsep}{18pt}
\begin{tabular}{l|c|c}
\hline
Method & PSNR (dB) & SSIM \\
\hline
SDE            & $28.25 \pm 11.20$ & $0.9345 \pm 0.0592$ \\
Mean Fill      & $18.76 \pm 5.33$ & $0.6876 \pm 0.0807$ \\
Telea          & $18.88 \pm 4.07$ & $0.7916 \pm 0.0422$ \\
Navier-Stokes  & $19.54 \pm 5.09$ & $0.7735 \pm 0.0684$ \\
\hline
\end{tabular}
\end{table}

\begin{table}[H]
\centering
\caption{Comparison on MNIST digit "4" (100 test samples)}
\label{tab:mnist_class_4}
\setlength{\tabcolsep}{18pt}
\begin{tabular}{l|c|c}
\hline
Method & PSNR (dB) & SSIM \\
\hline
SDE            & $18.06 \pm 5.27$ & $0.8317 \pm 0.0803$ \\
Mean Fill      & $15.60 \pm 2.71$ & $0.6832 \pm 0.0768$ \\
Telea          & $17.00 \pm 2.44$ & $0.7729 \pm 0.0726$ \\
Navier-Stokes  & $17.16 \pm 2.39$ & $0.7646 \pm 0.0706$ \\
\hline
\end{tabular}
\end{table}

\begin{table}[H]
\centering
\caption{Comparison on MNIST digit "8" (100 test samples)}
\label{tab:mnist_class_8}
\setlength{\tabcolsep}{18pt}
\begin{tabular}{l|c|c}
\hline
Method & PSNR (dB) & SSIM \\
\hline
SDE            & $17.68 \pm 5.28$ & $0.8407 \pm 0.0869$ \\
Mean Fill      & $15.25 \pm 2.46$ & $0.6965 \pm 0.0772$ \\
Telea          & $16.85 \pm 3.17$ & $0.7840 \pm 0.0753$ \\
Navier-Stokes  & $17.03 \pm 3.09$ & $0.7813 \pm 0.0757$ \\
\hline
\end{tabular}
\end{table}

\begin{figure}[H]
    \centering
    \includegraphics[width=1\textwidth]{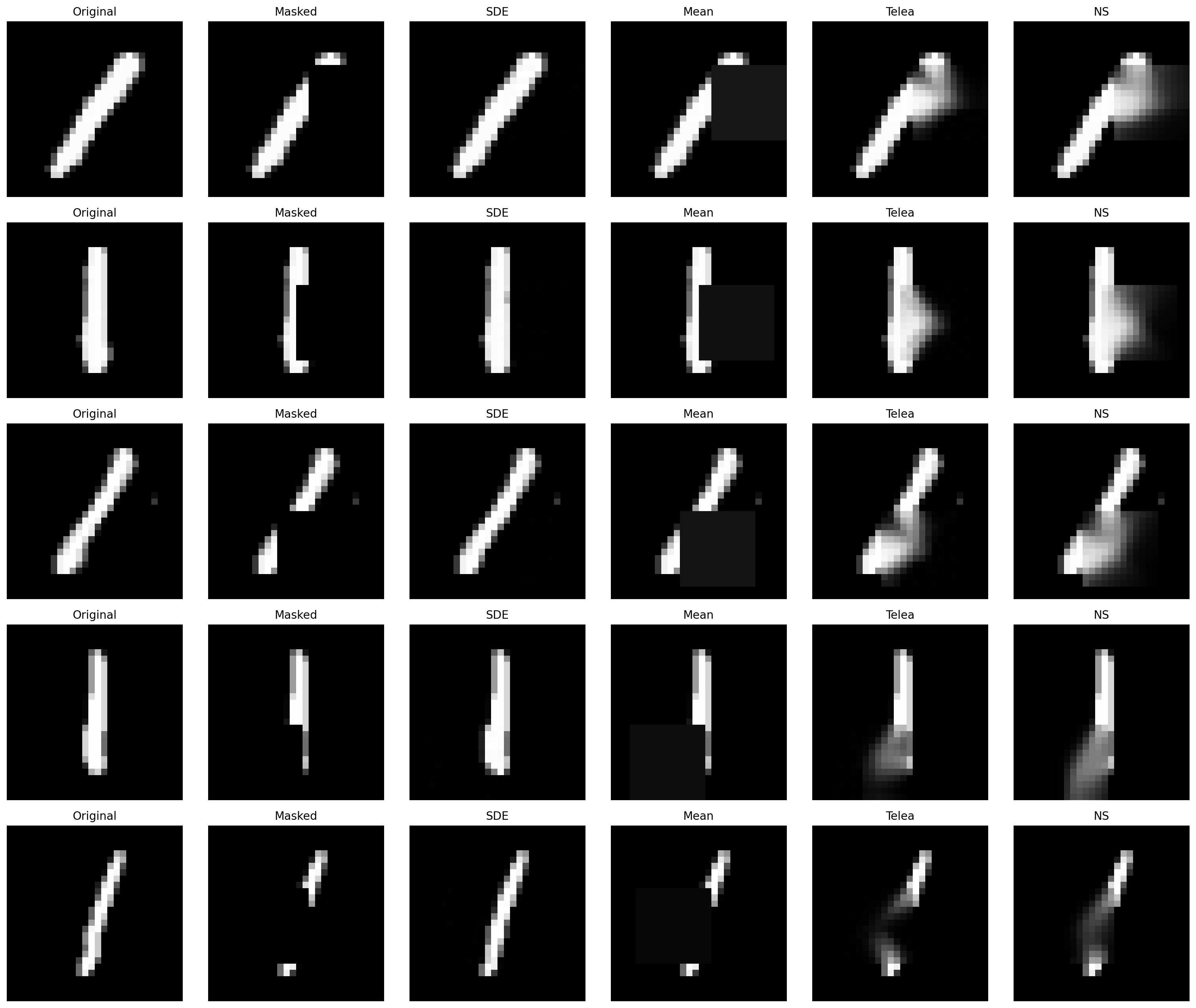}
    \caption{Label "1" Comparison of Image Inpainting Performance of SDE, Mean-Fill, Telea, and NS in MNIST}
    \label{mnist_1}
\end{figure}

\begin{figure}[H]
    \centering
    \includegraphics[width=1\textwidth]{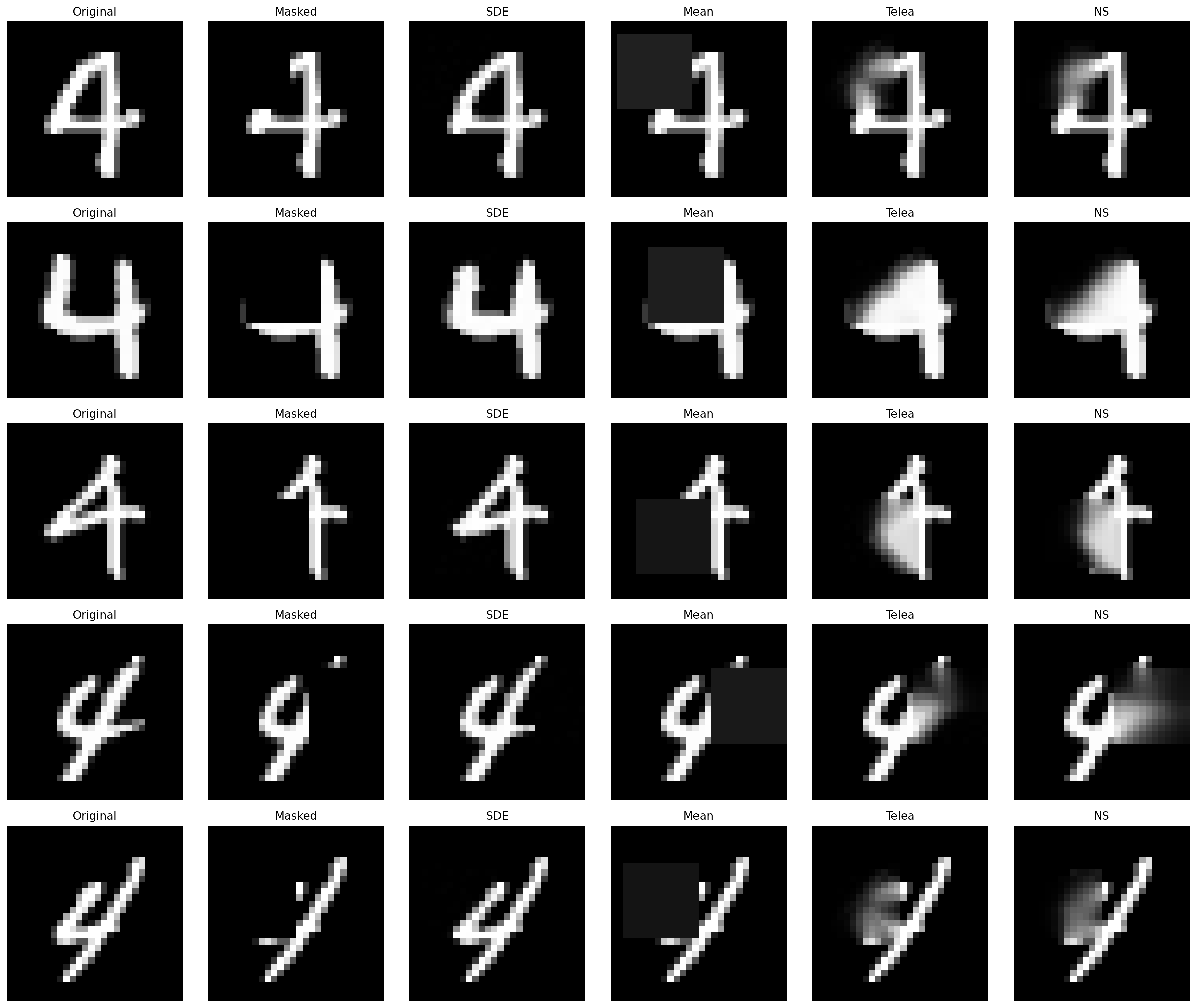}
    \caption{Label "4" Comparison of Image Inpainting Performance of SDE, Mean-Fill, Telea, and NS in MNIST}
    \label{mnist_4}
\end{figure}

\begin{figure}[H]
    \centering
    \includegraphics[width=1\textwidth]{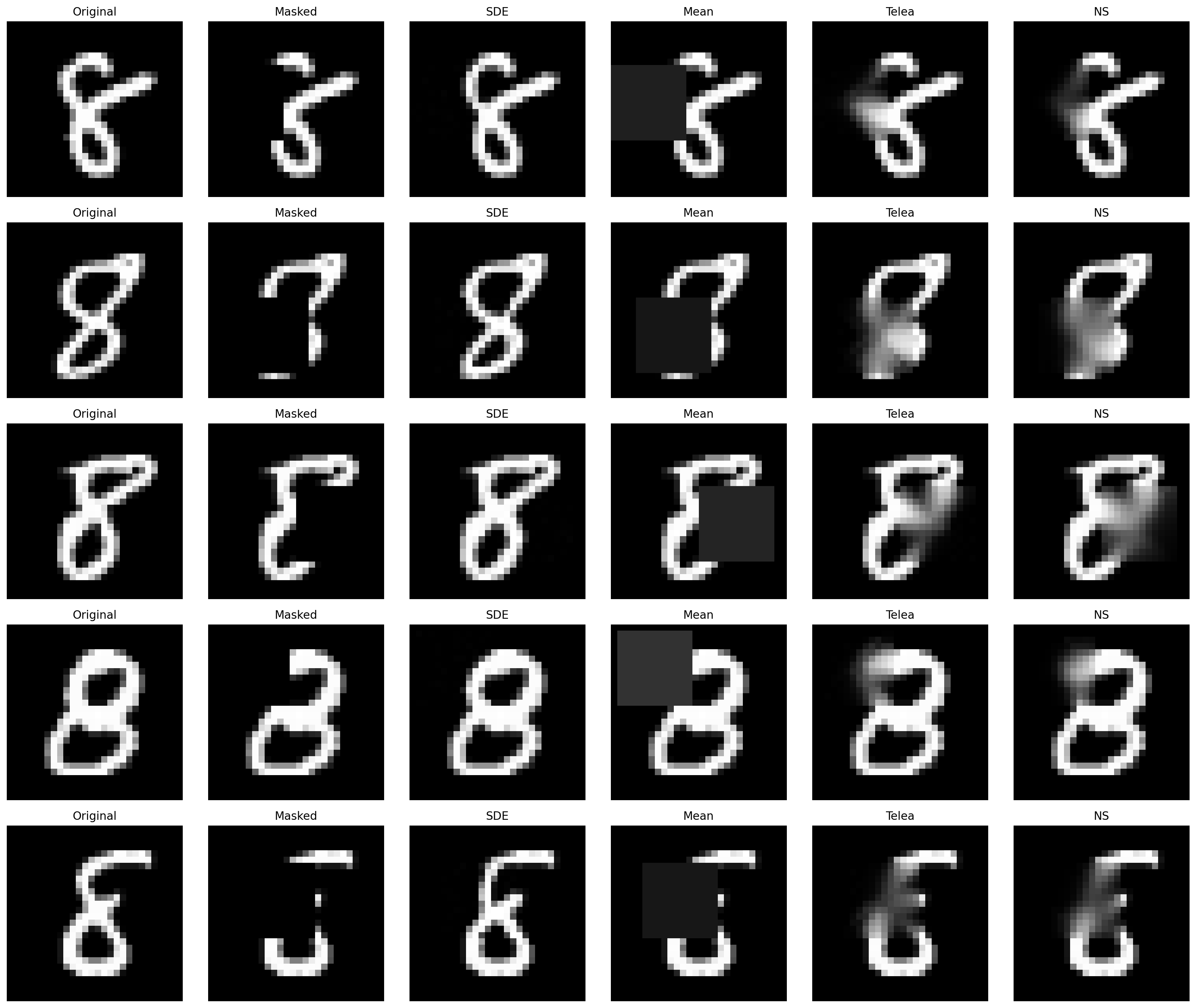}
    \caption{Label "8" Comparison of Image Inpainting Performance of SDE, Mean-Fill, Telea, and NS in MNIST}
    \label{mnist_8}
\end{figure}

These quantitative gains are consistent with the qualitative results in Figures~\ref{mnist_1},~\ref{mnist_4}, and~\ref{mnist_8}. Mean Fill leaves flat patches in the missing region, while Telea and Navier-Stokes often introduce blurring or geometric distortion, especially for thin strokes of "1" and closed contours of "8". In contrast, the proposed method better preserves vertical stroke continuity for "1", reconstructs the diagonal/cross structure of "4", and maintains the loop topology of "8".

We further evaluate class-specific priors on Fashion-MNIST, focusing on three representative categories: "Trouser", "Coat", and "Ankle boot". Tables~\ref{tab:fashion_class_trouser},~\ref{tab:fashion_class_coat}, and~\ref{tab:fashion_class_ankle_boot} show that the proposed method again outperforms all baselines. On "Trouser", it achieves a particularly strong result of $25.68$~dB PSNR and $0.9488$ SSIM. Similar improvements are observed on "Coat" and "Ankle boot", suggesting that category-level empirical priors are effective for recovering clothing silhouettes and structured object boundaries.

\begin{table}[H]
\centering
\caption{Comparison on Fashion-MNIST category "Trouser" (100 test samples)}
\label{tab:fashion_class_trouser}
\setlength{\tabcolsep}{18pt}
\begin{tabular}{l|c|c}
\hline
Method & PSNR (dB) & SSIM \\
\hline
SDE            & $25.68 \pm 4.24$ & $0.9488 \pm 0.0445$ \\
Mean Fill      & $15.60 \pm 2.14$ & $0.6879 \pm 0.0672$ \\
Telea          & $18.87 \pm 2.78$ & $0.8138 \pm 0.0564$ \\
Navier-Stokes  & $18.95 \pm 2.44$ & $0.8063 \pm 0.0569$ \\
\hline
\end{tabular}
\end{table}

\begin{table}[H]
\centering
\caption{Comparison on Fashion-MNIST category "Coat" (100 test samples)}
\label{tab:fashion_class_coat}
\setlength{\tabcolsep}{18pt}
\begin{tabular}{l|c|c}
\hline
Method & PSNR (dB) & SSIM \\
\hline
SDE            & $25.49 \pm 4.49$ & $0.8755 \pm 0.0565$ \\
Mean Fill      & $16.37 \pm 2.97$ & $0.7036 \pm 0.0571$ \\
Telea          & $20.79 \pm 4.33$ & $0.8306 \pm 0.0543$ \\
Navier-Stokes  & $21.07 \pm 4.34$ & $0.8336 \pm 0.0496$ \\
\hline
\end{tabular}
\end{table}

\begin{table}[H]
\centering
\caption{Comparison on Fashion-MNIST category "Ankle boot" (100 test samples)}
\label{tab:fashion_class_ankle_boot}
\setlength{\tabcolsep}{18pt}
\begin{tabular}{l|c|c}
\hline
Method & PSNR (dB) & SSIM \\
\hline
SDE            & $23.48 \pm 5.80$ & $0.8932 \pm 0.0567$ \\
Mean Fill      & $14.82 \pm 1.96$ & $0.6743 \pm 0.0688$ \\
Telea          & $19.32 \pm 3.04$ & $0.8309 \pm 0.0519$ \\
Navier-Stokes  & $19.94 \pm 3.50$ & $0.8444 \pm 0.0518$ \\
\hline
\end{tabular}
\end{table}

\begin{figure}[H]
    \centering
    \includegraphics[width=1\textwidth]{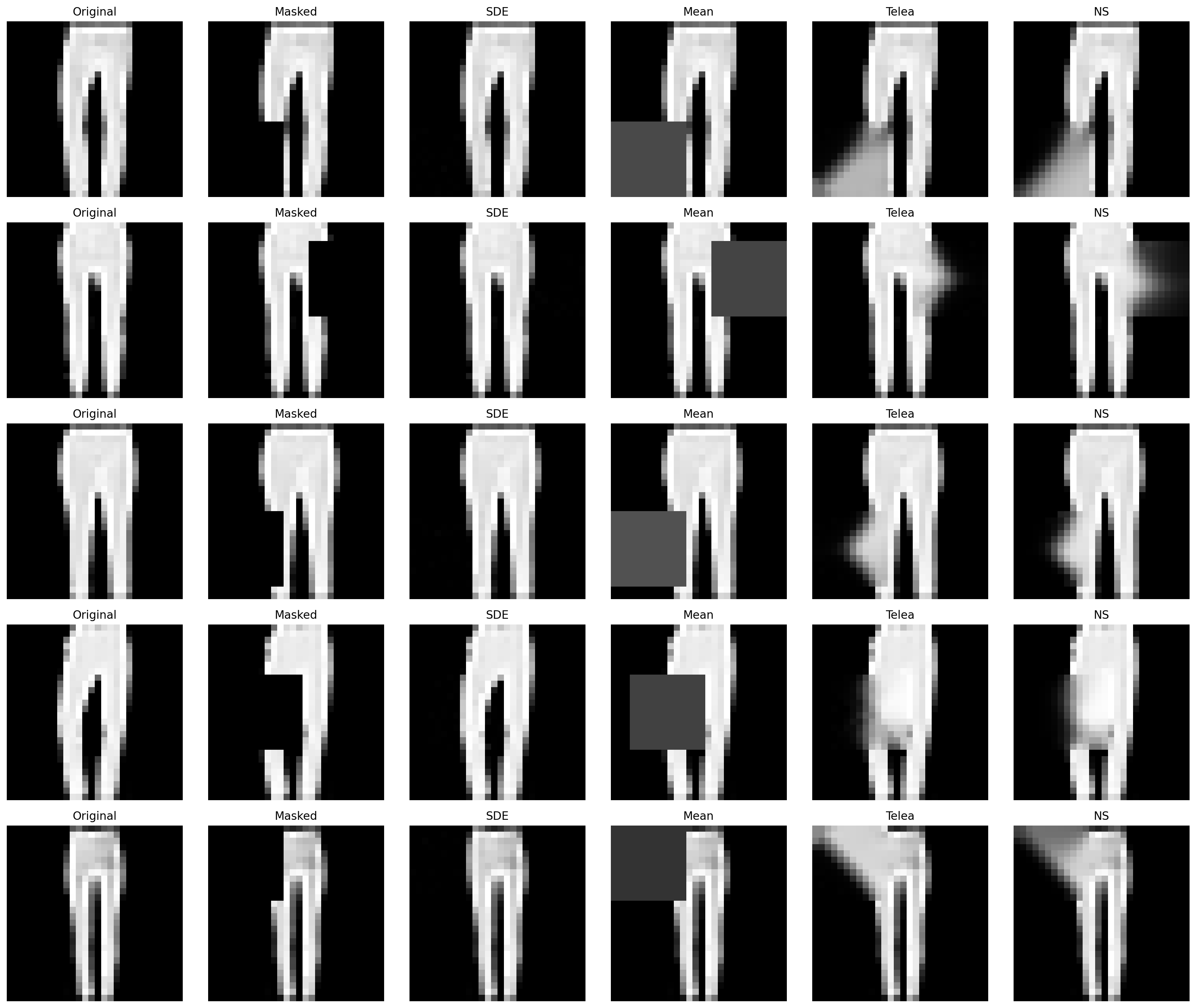}
    \caption{Label "Trouser" Comparison of Image Inpainting Performance of SDE, Mean-Fill, Telea, and NS in Fashion-MNIST}
    \label{fashion_1}
\end{figure}

\begin{figure}[H]
    \centering
    \includegraphics[width=1\textwidth]{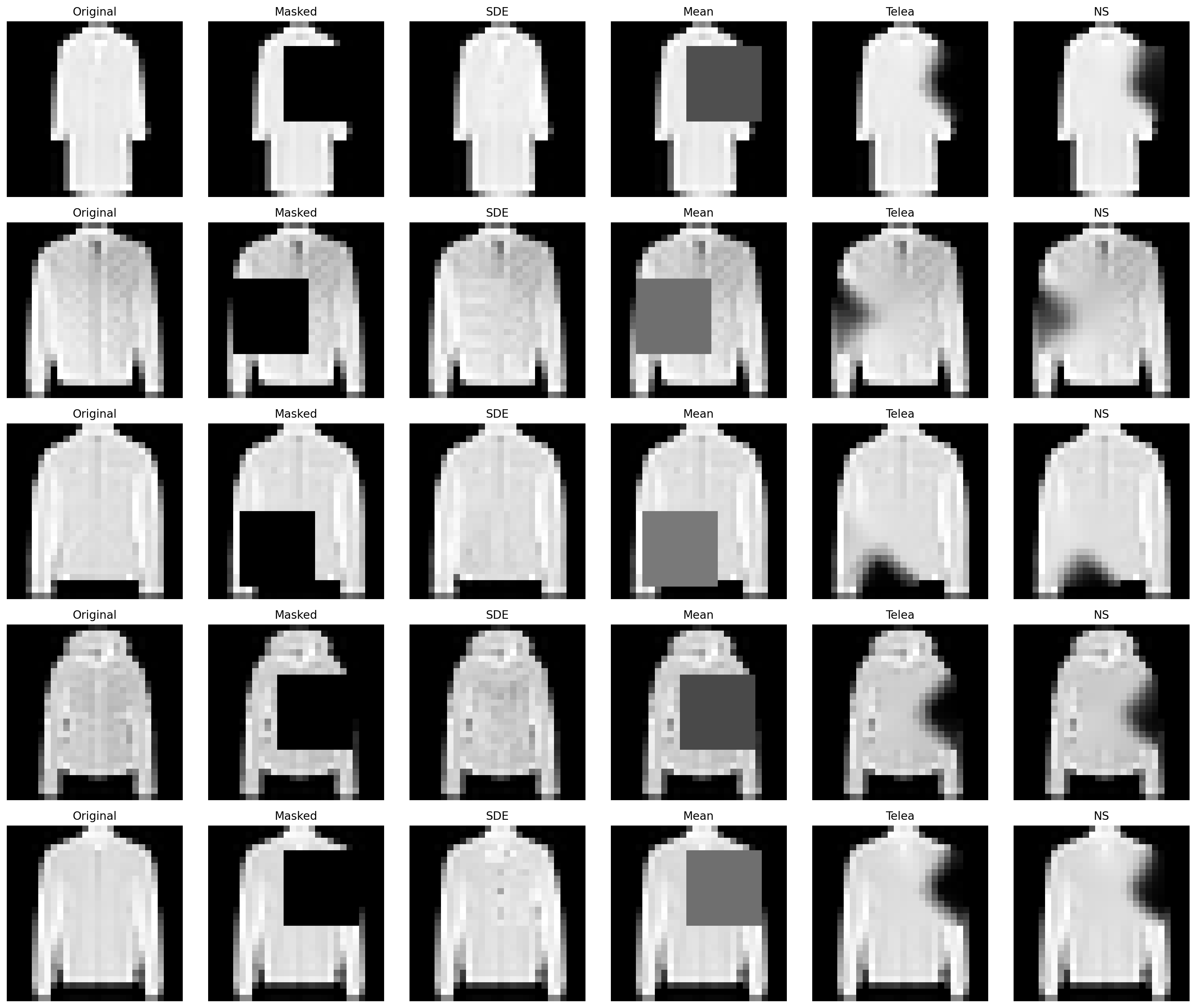}
    \caption{Label "Coat" Comparison of Image Inpainting Performance of SDE, Mean-Fill, Telea, and NS in Fashion-MNIST}
    \label{fashion_4}
\end{figure}

\begin{figure}[H]
    \centering
    \includegraphics[width=1\textwidth]{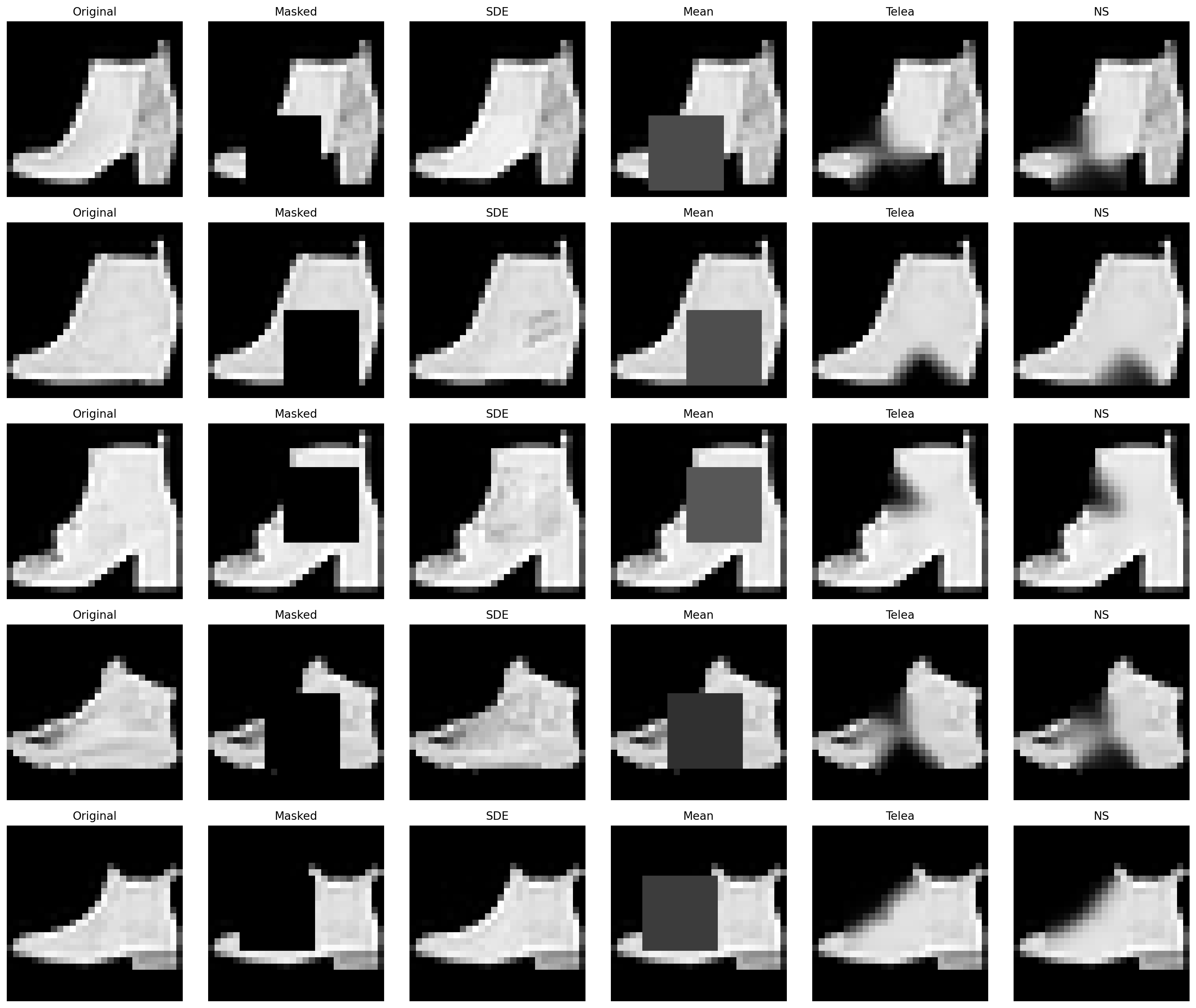}
    \caption{Label "Ankle boot" Comparison of Image Inpainting Performance of SDE, Mean-Fill, Telea, and NS in Fashion-MNIST}
    \label{fashion_9}
\end{figure}

Figures~\ref{fashion_1},~\ref{fashion_4}, and~\ref{fashion_9} provide corresponding qualitative comparisons. Mean Fill fails to recover semantic content, while Telea and Navier-Stokes tend to blur edges and distort geometry, especially around "trouser" legs, "coat" boundaries, and "boot" heels. By contrast, the SDE method better preserves object symmetry and global layout, such as the two-leg structure of trousers and the sharp contour of "ankle boots".

Additional class-specific results, including quantitative tables and qualitative examples for MNIST digits "0", "2", and "5", as well as Fashion-MNIST categories "T-shirt/top", "Bag", and "Sandal", are provided in Appendix~\ref{app:class_specific}. These supplementary results further support the robustness of class-specific empirical priors across different object structures.

\subsection{Experiments on Industrial Images}

We evaluate the proposed method on two MVTec AD categories, Bottle and Capsule. All images are resized to $128\times128$, with a $32\times32$ missing region. We use $\nu=0.05$, $\beta=2.0$, $T=1.0$, and a local context kernel of size $5$.

Tables~\ref{tab:mvtec_bottle} and~\ref{tab:mvtec_capsule} report the average PSNR and SSIM. On both categories, the proposed SDE method substantially outperforms Mean Fill, Telea, and Navier--Stokes. In particular, it improves PSNR by about $8.48$~dB on Bottle and about $12.94$~dB on Capsule relative to the stronger PDE baseline. The corresponding SSIM values are also consistently higher, indicating more faithful recovery of industrial structures and textures.

\begin{table}[t]
\centering
\caption{Comparison Report of Image Inpainting Methods on MVTEC-Bottle (100 test samples)}
\label{tab:mvtec_bottle}
\begin{tabular}{lcc}
\toprule
Method & PSNR (dB) & SSIM \\
\midrule
SDE & $39.61\pm5.00$ & $0.9849\pm0.0104$ \\
Mean Fill  & $21.85\pm1.15$ & $0.9293\pm0.0055$ \\
Telea      & $30.93\pm6.21$ & $0.9642\pm0.0114$ \\
Navier-Stokes & $31.13\pm6.23$ & $0.9645\pm0.0118$ \\
\bottomrule
\end{tabular}
\end{table}
\begin{table}[t]
\centering
\caption{Comparison Report of Image Inpainting Methods on MVTEC-Capsule (100 test samples)}
\label{tab:mvtec_capsule}
\begin{tabular}{lcc}
\toprule
Method & PSNR (dB) & SSIM \\
\midrule
SDE & $46.63\pm3.67$ & $0.9920\pm0.0036$ \\
Mean Fill  & $23.68\pm4.28$ & $0.9428\pm0.0161$ \\
Telea      & $33.54\pm7.25$ & $0.9782\pm0.0120$ \\
Navier-Stokes & $33.69\pm7.24$ & $0.9789\pm0.0122$ \\
\bottomrule
\end{tabular}
\end{table}

\begin{figure}[H]
    \centering
    \includegraphics[width=1\textwidth]{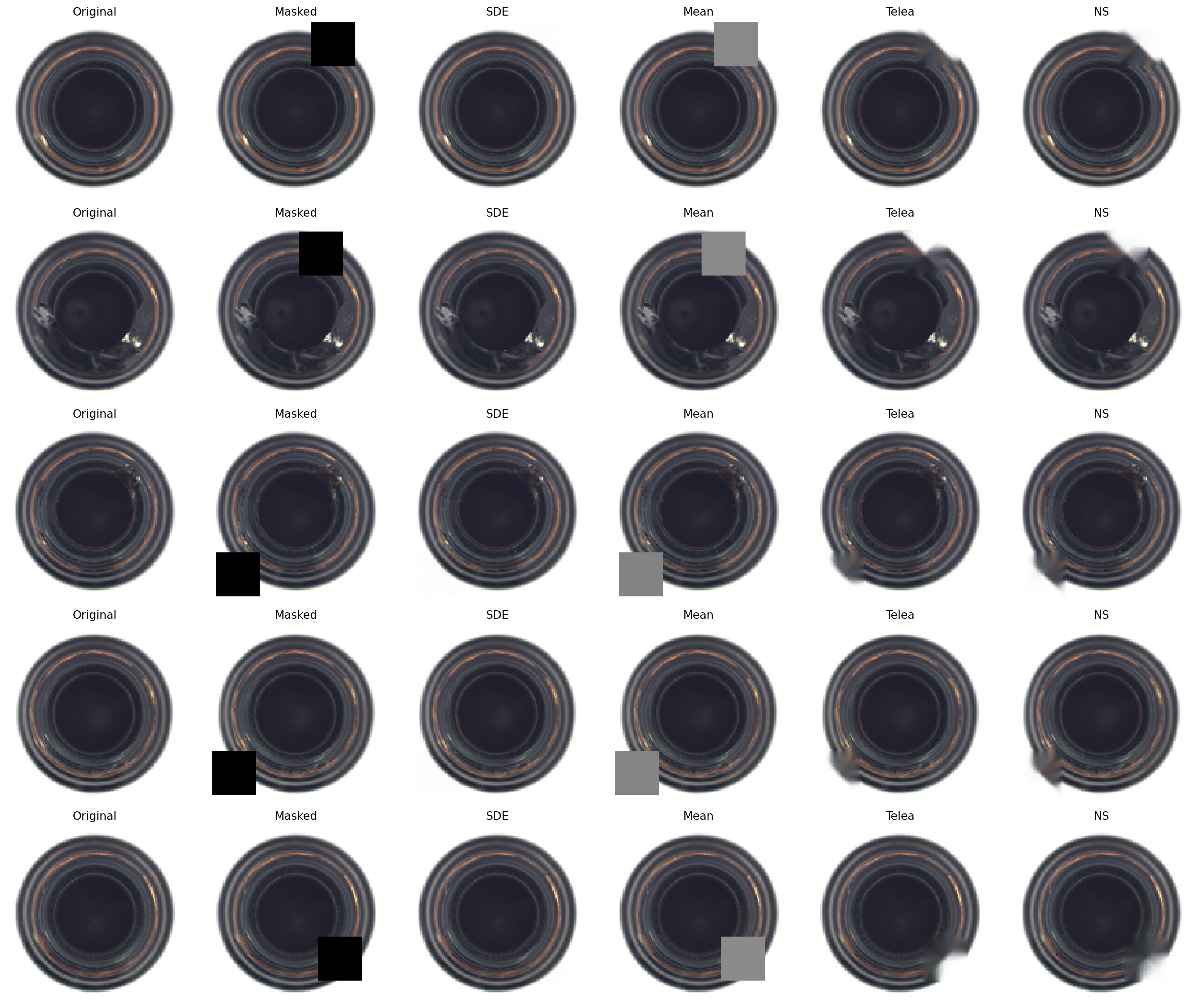}
    \caption{Label "Bottle" Comparison of Image Inpainting Performance of SDE, Mean-Fill, Telea, and NS in MVTEC}
    \label{mvtec_bottle}
\end{figure}

\begin{figure}[H]
    \centering
    \includegraphics[width=1\textwidth]{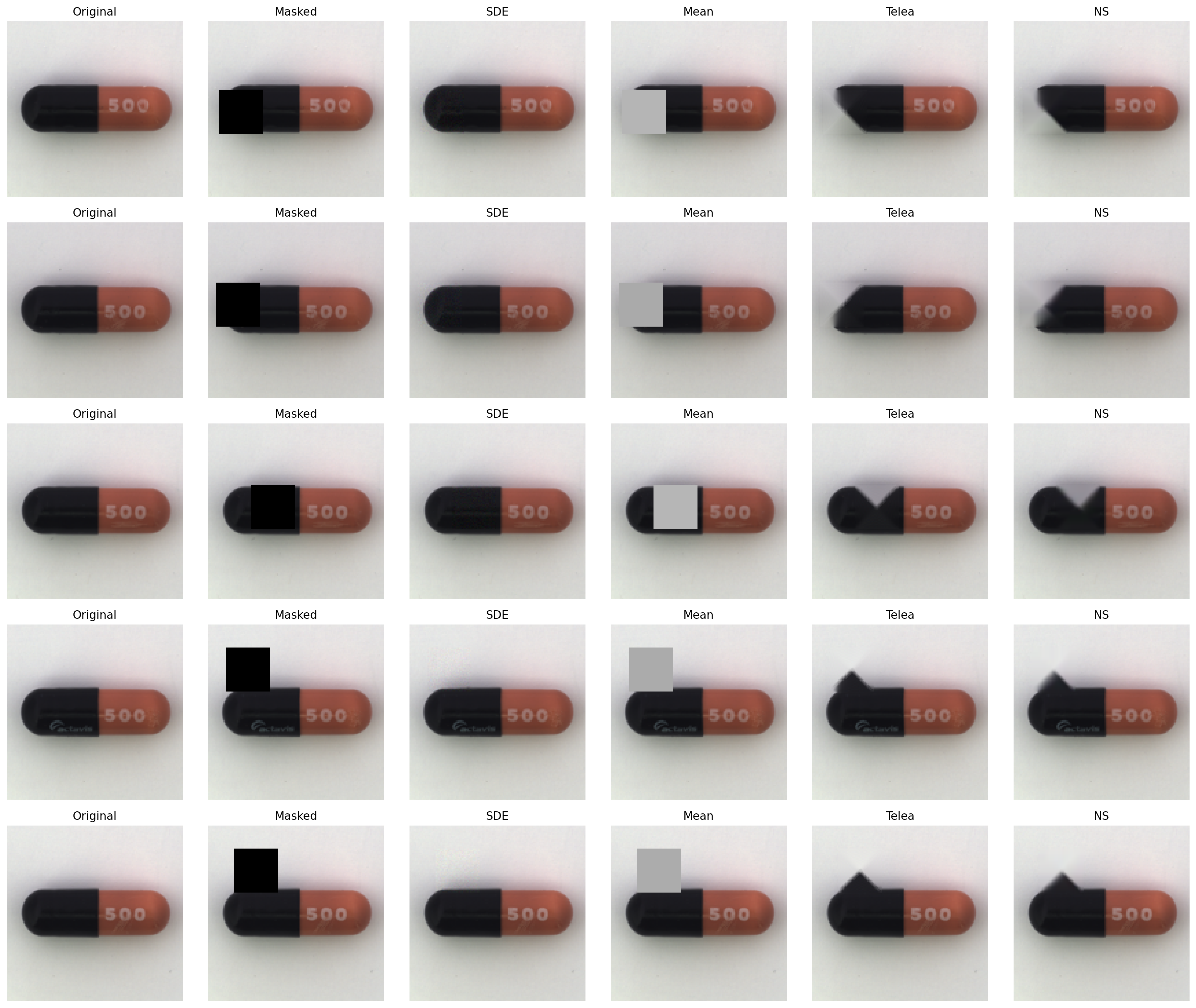}
    \caption{Label "Capsule" Comparison of Image Inpainting Performance of SDE, Mean-Fill, Telea, and NS in MVTEC}
    \label{mvtec_capsule}
\end{figure}

Figures~\ref{mvtec_bottle} and~\ref{mvtec_capsule} provide qualitative comparisons. The Mean Fill still leaves a flat patch and Telea/Navier--Stokes tend to blur object boundaries and reflective patterns, but our SDE method restores sharp contours and local appearance that remain close to the original images. These results suggest that an empirical reference prior is particularly effective for industrial images with stable object geometry and limited intra-class variation.

\subsection{Extension to CelebA Face Images}

To further evaluate the applicability of the proposed training-free SDE-based inpainting method beyond grayscale and industrial images, we conduct experiments on the CelebA face dataset. All images are loaded in RGB format, resized to $64\times64$ and a random $16\times16$ square mask is used. We set the context kernel size to $5$, the candidate pool size to 4096, and the subset size to 1024. This low-resolution setting is used as a controlled evaluation protocol due to the computational cost of non-parametric sample matching in high-dimensional image spaces.

For each test image, we randomly place a $16 \times 16$ square mask. Missing pixels are initialized with the normalized lower-bound value $-1$, while observed pixels are clamped to their original values throughout the SDE iterations. To improve sample matching in the high-dimensional RGB setting, kernel weights are computed using a local context region around the missing block rather than the entire image. Specifically, we dilate the missing mask with a $5 \times 5$ kernel and use the observed pixels in this local ring as the context. We first select the top 4096 candidate reference samples according to the context distance, and at each SDE step a subset of 1024 samples is used to estimate the empirical correction.

We compare the proposed method with Mean Fill, Telea's fast marching inpainting method, and the Navier-Stokes inpainting method implemented in OpenCV. Following the non-parametric nature of our method, no neural network training is performed.

Table~\ref{tab:celeba_100} reports the quantitative results over 100 randomly selected test samples. The proposed SDE method achieves a PSNR of $29.81$~dB and an SSIM of $0.9590$, outperforming Mean Fill as well as Telea and Navier--Stokes under this setting. This suggests that, for face-structure occlusions, an empirical reference prior can be more effective than purely local PDE propagation, while still requiring no network training.

\begin{figure}[H]
    \centering
    \includegraphics[width=1\textwidth]{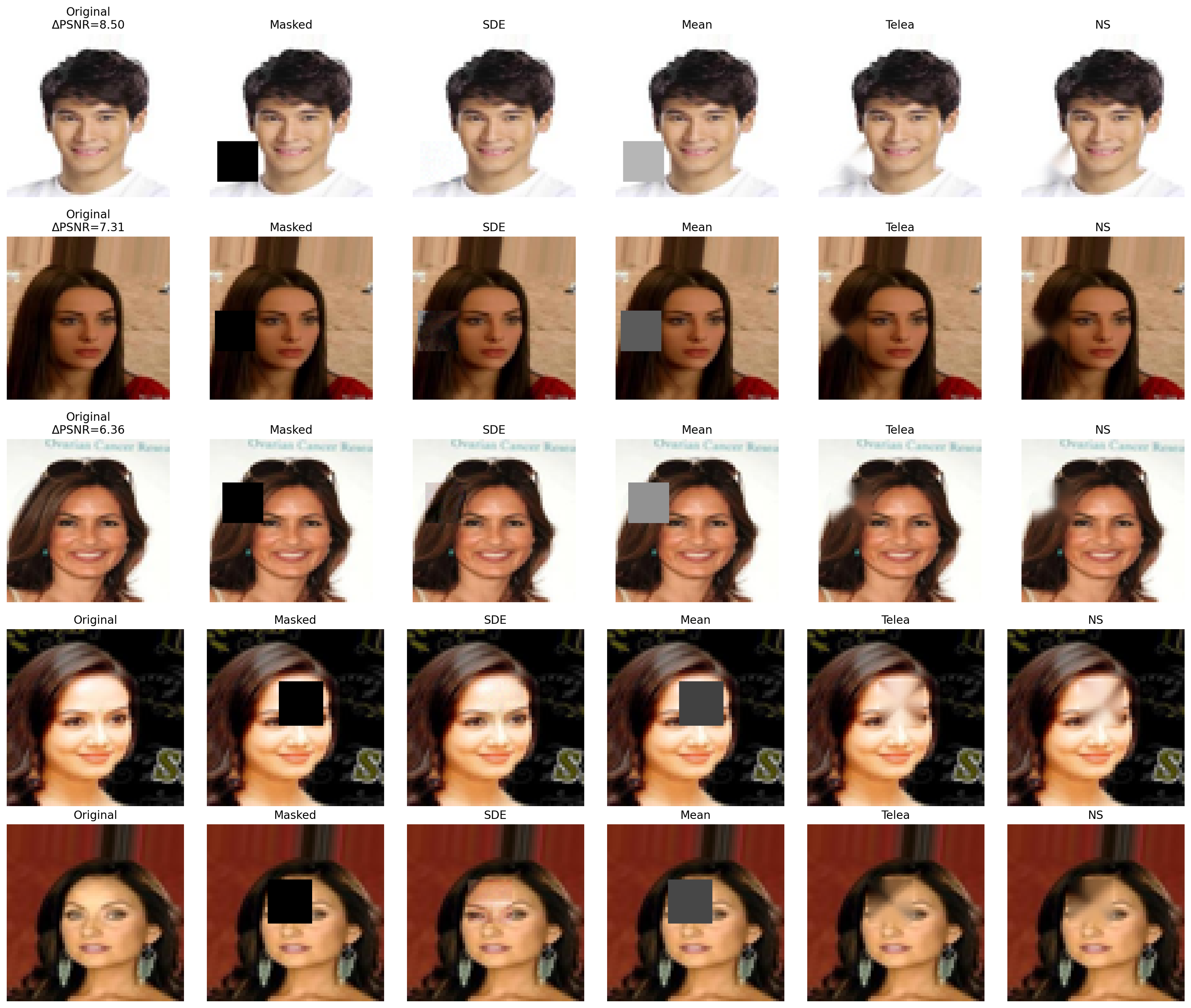}
    \caption{Comparison of Image Inpainting Performance of SDE, Mean-Fill, Telea, and NS in CelebA}
    \label{celeba}
\end{figure}

\begin{figure}[H]
    \centering
    \includegraphics[width=1\textwidth]{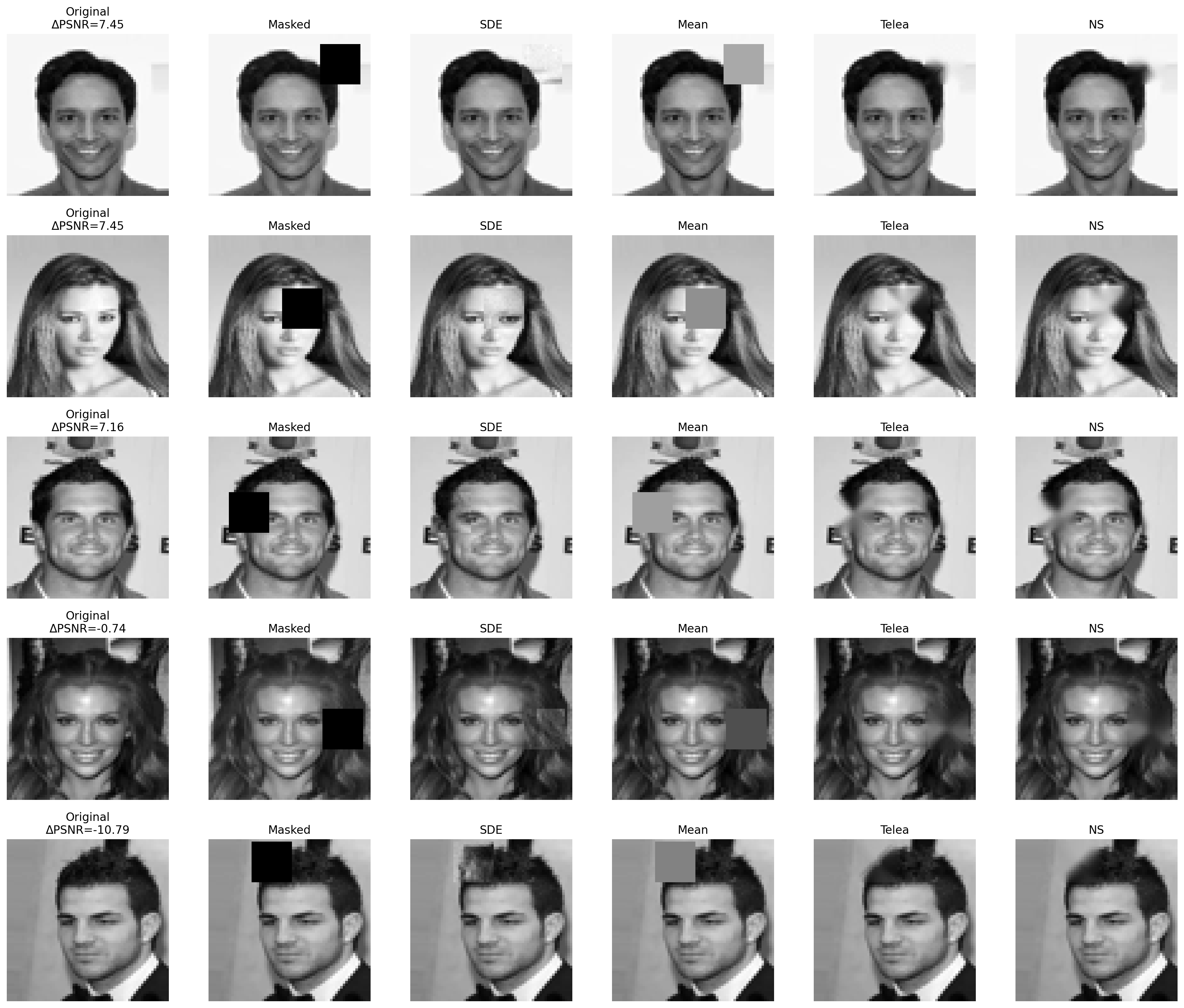}
    \caption{Comparison of Image Inpainting Performance of SDE, Mean-Fill, Telea, and NS on grayscale CelebA}
    \label{celeba_gray}
\end{figure}

Qualitative examples are shown in Figure~\ref{celeba}. Telea and Navier-Stokes may introduce local propagation artifacts when the occlusion overlaps facial structures, such as dragging nearby hair, background, or boundary textures into the missing region. In contrast, the proposed SDE method more often preserves facial appearance by leveraging empirical face priors from the reference dataset. These observations are consistent with the quantitative results in Table~\ref{tab:celeba_100}. Figure~\ref{celeba_gray} additionally presents qualitative results on grayscale CelebA images, where the proposed method also produces visually coherent facial completions.

\begin{table}[H]
\centering
\caption{Comparison of Image Inpainting Methods on CelebA (100 test samples)}
\label{tab:celeba_100}
\setlength{\tabcolsep}{18pt}
\begin{tabular}{l|c|c}
\hline
Method & PSNR (dB) & SSIM \\
\hline
SDE            & $\mathbf{29.81 \pm 3.22}$ & $\mathbf{0.9590 \pm 0.0150}$ \\
Mean Fill      & $23.77 \pm 2.72$ & $0.9183 \pm 0.0146$ \\
Telea          & $28.62 \pm 3.11$ & $0.9495 \pm 0.0131$ \\
Navier-Stokes  & $28.58 \pm 3.13$ & $0.9492 \pm 0.0130$ \\
\hline
\end{tabular}
\end{table}

\subsection{Ablation Study}

We examine the effect of the diffusion coefficient $\nu$, which controls the stochastic intensity of the reverse dynamics and also influences the scale $\rho(t)$ of the empirical correction. All other settings are kept fixed, and $\nu$ is varied over $\{0.05,0.10,0.30,1.00\}$.

\begin{table}[H]
\centering
\caption{Ablation study of the diffusion coefficient $\nu$ on MNIST (100 test samples)}
\label{tab:ablation_nu_mnist}
\setlength{\tabcolsep}{18pt}
\begin{tabular}{c|c|c}
\hline
$\nu$ & PSNR (dB) & SSIM \\
\hline
0.05 & $19.46 \pm 7.26$ & $0.8582 \pm 0.0994$ \\
0.10 & $19.40 \pm 6.97$ & $0.8571 \pm 0.0984$ \\
0.30 & $19.30 \pm 6.52$ & $0.8532 \pm 0.0951$ \\
1.00 & $19.20 \pm 6.05$ & $0.8442 \pm 0.0889$ \\
\hline
\end{tabular}
\end{table}

\begin{table}[H]
\centering
\caption{Ablation study of the diffusion coefficient $\nu$ on MNIST (500 test samples)}
\label{tab:ablation_nu_mnist_500}
\setlength{\tabcolsep}{18pt}
\begin{tabular}{c|c|c}
\hline
$\nu$ & PSNR (dB) & SSIM \\
\hline
0.05 & $18.83 \pm 6.93$ & $0.8429 \pm 0.0979$ \\
0.10 & $18.78 \pm 6.66$ & $0.8420 \pm 0.0972$ \\
0.30 & $18.71 \pm 6.23$ & $0.8386 \pm 0.0948$ \\
1.00 & $18.63 \pm 5.76$ & $0.8302 \pm 0.0900$ \\
\hline
\end{tabular}
\end{table}

\begin{table}[H]
\centering
\caption{Ablation study of the diffusion coefficient $\nu$ on Fashion-MNIST (100 test samples)}
\label{tab:ablation_nu_fashion_mnist_100}
\setlength{\tabcolsep}{18pt}
\begin{tabular}{c|c|c}
\hline
$\nu$ & PSNR (dB) & SSIM \\
\hline
0.05 & $23.61 \pm 5.20$ & $0.8726 \pm 0.0773$ \\
0.10 & $23.60 \pm 5.18$ & $0.8719 \pm 0.0767$ \\
0.30 & $23.58 \pm 5.09$ & $0.8693 \pm 0.0747$ \\
1.00 & $23.50 \pm 4.89$ & $0.8627 \pm 0.0710$ \\
\hline
\end{tabular}
\end{table}

\begin{table}[H]
\centering
\caption{Ablation study of the diffusion coefficient $\nu$ on Fashion-MNIST (500 test samples)}
\label{tab:ablation_nu_fashion_mnist_500}
\setlength{\tabcolsep}{18pt}
\begin{tabular}{c|c|c}
\hline
$\nu$ & PSNR (dB) & SSIM \\
\hline
0.05 & $23.63 \pm 5.43$ & $0.8739 \pm 0.0741$ \\
0.10 & $23.61 \pm 5.34$ & $0.8732 \pm 0.0735$ \\
0.30 & $23.57 \pm 5.16$ & $0.8705 \pm 0.0718$ \\
1.00 & $23.48 \pm 4.89$ & $0.8636 \pm 0.0679$ \\
\hline
\end{tabular}
\end{table}

Tables~\ref{tab:ablation_nu_mnist}--\ref{tab:ablation_nu_fashion_mnist_500} show a consistent but moderate trend: decreasing $\nu$ improves both PSNR and SSIM on MNIST and Fashion-MNIST. The same ordering is observed for both 100 and 500 test samples, indicating that the result is stable with respect to the evaluation size. Among the tested values, $\nu=0.05$ achieves the best average performance and is therefore used in the remaining experiments.

\section{Conclusion}
\label{conclusion}

In this work, we developed a non-parametric image inpainting method based on data-guided stochastic dynamics. By exploiting empirical
statistics from a reference dataset, the proposed approach guides the evolution of a masked image toward high-density regions of the data distribution without training a parametric model or explicitly fitting a parametric density.

Extensive experiments on MNIST and Fashion-MNIST demonstrate that the proposed method consistently outperforms Mean Fill, Telea, and
Navier--Stokes inpainting in terms of PSNR, SSIM, and visual quality. The reconstructed images exhibit improved class-consistent structures, including digit strokes and garment contours. The performance gains are even more pronounced on the Bottle and Capsule categories of MVTec AD, indicating that empirical reference priors are particularly effective for industrial images with stable geometry and limited intra-class variation. On CelebA, the proposed method also achieves higher average
PSNR and SSIM than the considered classical baselines under our experimental setting, while the qualitative results show improved preservation of facial structures for several occlusion patterns.

The current method is most effective on structured or category-consistent image collections, such as handwritten digits, fashion items, and industrial objects. Its advantage becomes more moderate on heterogeneous natural-image datasets with substantial variation in appearance, pose, and background, as illustrated by the CelebA experiments. In addition, the computational cost of empirical sample matching may increase with image resolution and reference-set
size.

Future work will focus on improving the scalability and robustness of the proposed framework. Possible directions include more efficient reference-sample retrieval, feature-space rather than pixel-space guidance, extensions to irregular and larger missing regions, and evaluation on higher-resolution and more diverse image datasets. These developments may broaden the applicability of the method while preserving its non-parametric and training-free stochastic formulation.

\bibliographystyle{apacite}
\bibliography{bibliography.bib}

\appendix
\section{Additional Class-specific Results}
\label{app:class_specific}

\begin{table}[H]
\centering
\caption{Comparison on MNIST digit "0" (100 test samples)}
\label{tab:mnist_class_0}
\setlength{\tabcolsep}{18pt}
\begin{tabular}{l|c|c}
\hline
Method & PSNR (dB) & SSIM \\
\hline
SDE            & $19.02 \pm 3.57$ & $0.8945 \pm 0.0608$ \\
Mean Fill      & $14.95 \pm 1.87$ & $0.7167 \pm 0.0636$ \\
Telea          & $16.25 \pm 2.63$ & $0.7860 \pm 0.0899$ \\
Navier-Stokes  & $16.53 \pm 2.61$ & $0.7918 \pm 0.0850$ \\
\hline
\end{tabular}
\end{table}

\begin{table}[H]
\centering
\caption{Comparison on MNIST digit “2” (100 test samples)}
\label{tab:mnist_class_2}
\setlength{\tabcolsep}{18pt}
\begin{tabular}{l|c|c}
\hline
Method & PSNR (dB) & SSIM \\
\hline
SDE            & $17.41 \pm 3.47$ & $0.8399 \pm 0.0832$ \\
Mean Fill      & $15.13 \pm 2.43$ & $0.6982 \pm 0.0703$ \\
Telea          & $16.59 \pm 2.40$ & $0.7840 \pm 0.0735$ \\
Navier-Stokes  & $16.89 \pm 2.49$ & $0.7835 \pm 0.0694$ \\
\hline
\end{tabular}
\end{table}

\begin{table}[H]
\centering
\caption{Comparison on MNIST digit "5" (100 test samples)}
\label{tab:mnist_class_5}
\setlength{\tabcolsep}{18pt}
\begin{tabular}{l|c|c}
\hline
Method & PSNR (dB) & SSIM \\
\hline
SDE            & $17.74 \pm 3.30$ & $0.8434 \pm 0.0832$ \\
Mean Fill      & $15.34 \pm 2.27$ & $0.6908 \pm 0.0780$ \\
Telea          & $16.76 \pm 2.59$ & $0.7777 \pm 0.0796$ \\
Navier-Stokes  & $16.69 \pm 2.35$ & $0.7655 \pm 0.0787$ \\
\hline
\end{tabular}
\end{table}

\begin{figure}[H]
    \centering
    \includegraphics[width=1\textwidth]{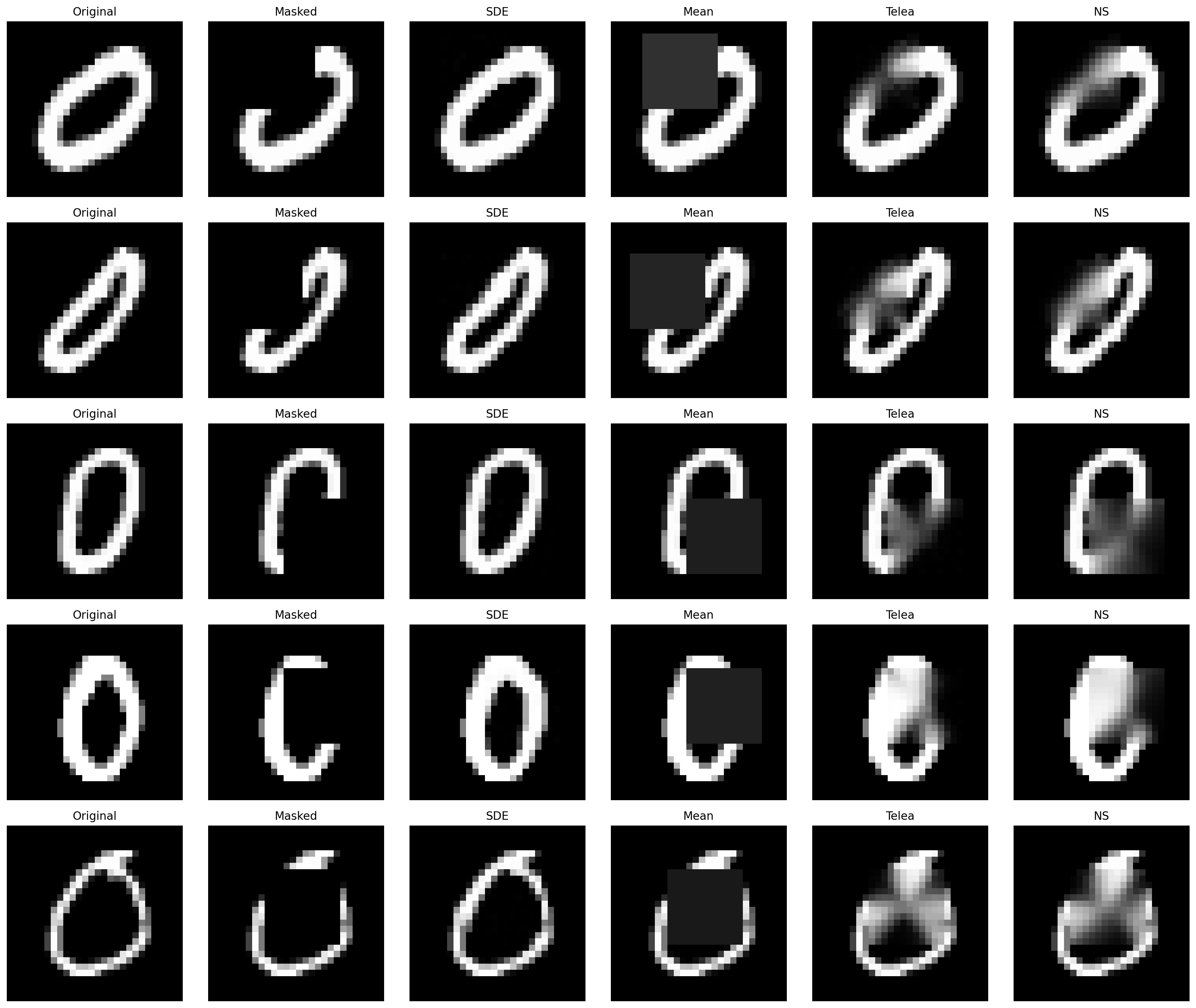}
    \caption{Label "0" Comparison of Image Inpainting Performance of SDE, Mean-Fill, Telea, and NS in MNIST}
    \label{mnist_0}
\end{figure}

\begin{figure}[H]
    \centering
    \includegraphics[width=1\textwidth]{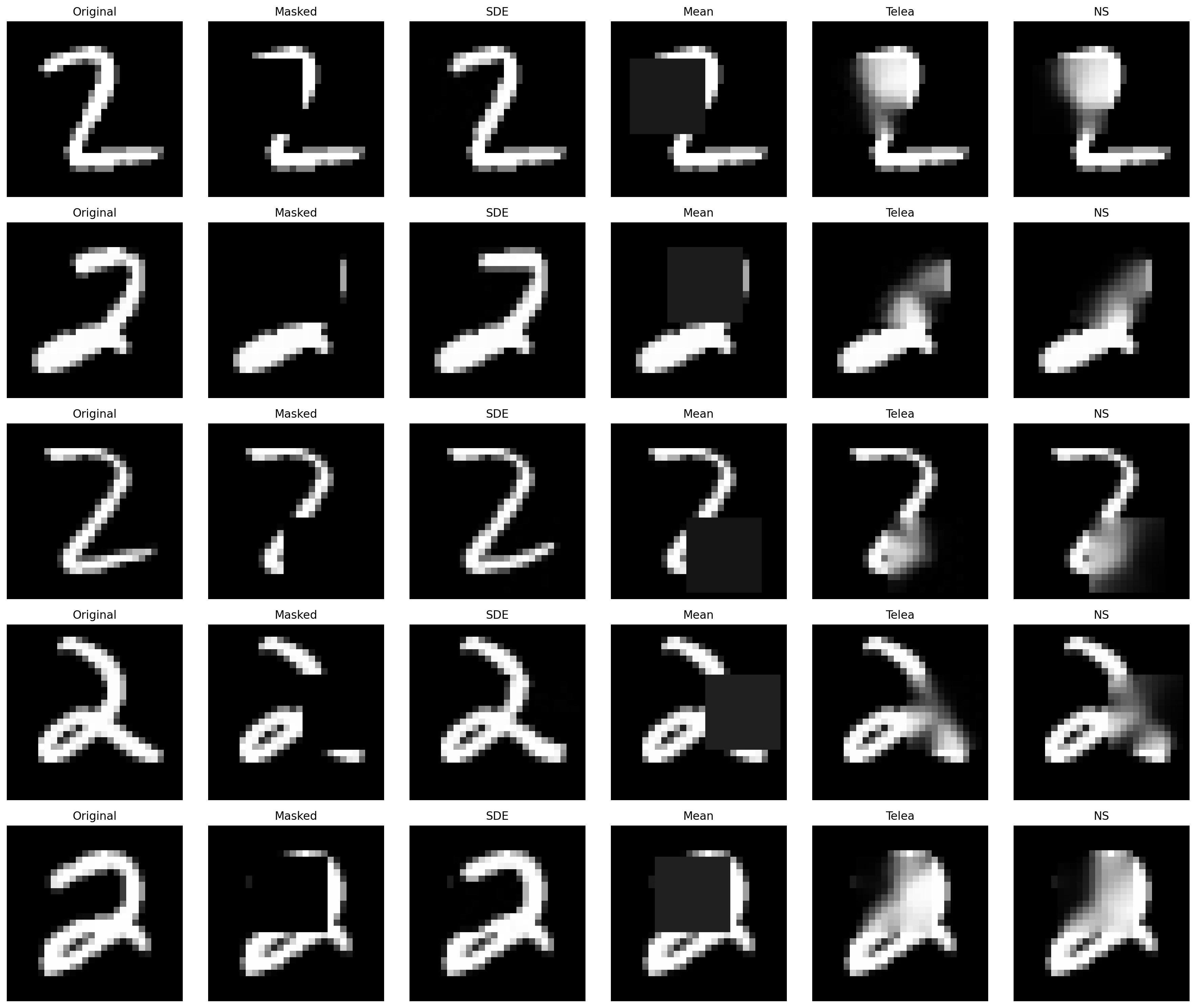}
    \caption{Label "2" Comparison of Image Inpainting Performance of SDE, Mean-Fill, Telea, and NS in MNIST}
    \label{mnist_2}
\end{figure}

\begin{figure}[H]
    \centering
    \includegraphics[width=1\textwidth]{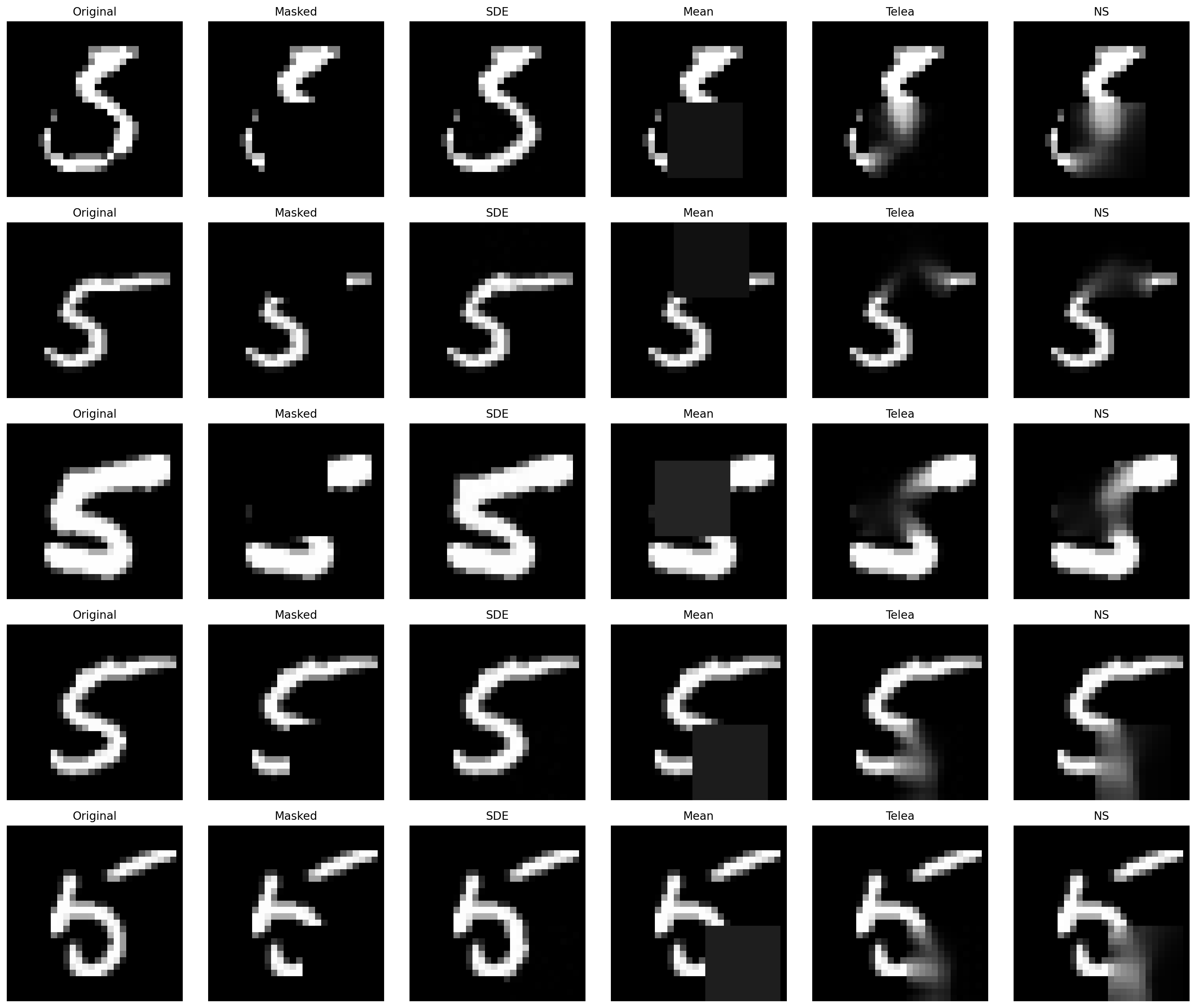}
    \caption{Label "5" Comparison of Image Inpainting Performance of SDE, Mean-Fill, Telea, and NS in MNIST}
    \label{mnist_5}
\end{figure}

\begin{table}[H]
\centering
\caption{Comparison on Fashion-MNIST category "T-shirt/top" (100 test samples)}
\label{tab:fashion_class_tshirt}
\setlength{\tabcolsep}{18pt}
\begin{tabular}{l|c|c}
\hline
Method & PSNR (dB) & SSIM \\
\hline
SDE            & $24.67 \pm 5.27$ & $0.8658 \pm 0.0680$ \\
Mean Fill      & $17.75 \pm 3.39$ & $0.7185 \pm 0.0673$ \\
Telea          & $21.78 \pm 3.76$ & $0.8362 \pm 0.0558$ \\
Navier-Stokes  & $21.87 \pm 3.53$ & $0.8384 \pm 0.0508$ \\
\hline
\end{tabular}
\end{table}

\begin{table}[H]
\centering
\caption{Comparison on Fashion-MNIST category "Bag" (100 test samples)}
\label{tab:fashion_class_bag}
\setlength{\tabcolsep}{18pt}
\begin{tabular}{l|c|c}
\hline
Method & PSNR (dB) & SSIM \\
\hline
SDE            & $22.05 \pm 4.73$ & $0.8326 \pm 0.0667$ \\
Mean Fill      & $16.19 \pm 2.25$ & $0.6961 \pm 0.0561$ \\
Telea          & $20.51 \pm 4.34$ & $0.8317 \pm 0.0530$ \\
Navier-Stokes  & $20.58 \pm 4.32$ & $0.8315 \pm 0.0523$ \\
\hline
\end{tabular}
\end{table}

\begin{table}[H]
\centering
\caption{Comparison on Fashion-MNIST category "Sandal" (100 test samples)}
\label{tab:fashion_class_sandal}
\setlength{\tabcolsep}{18pt}
\begin{tabular}{l|c|c}
\hline
Method & PSNR (dB) & SSIM \\
\hline
SDE            & $20.92 \pm 5.08$ & $0.8488 \pm 0.0800$ \\
Mean Fill      & $18.41 \pm 3.25$ & $0.6981 \pm 0.0706$ \\
Telea          & $20.05 \pm 3.76$ & $0.8006 \pm 0.0578$ \\
Navier-Stokes  & $20.08 \pm 3.61$ & $0.7970 \pm 0.0597$ \\
\hline
\end{tabular}
\end{table}

\begin{figure}[H]
    \centering
    \includegraphics[width=1\textwidth]{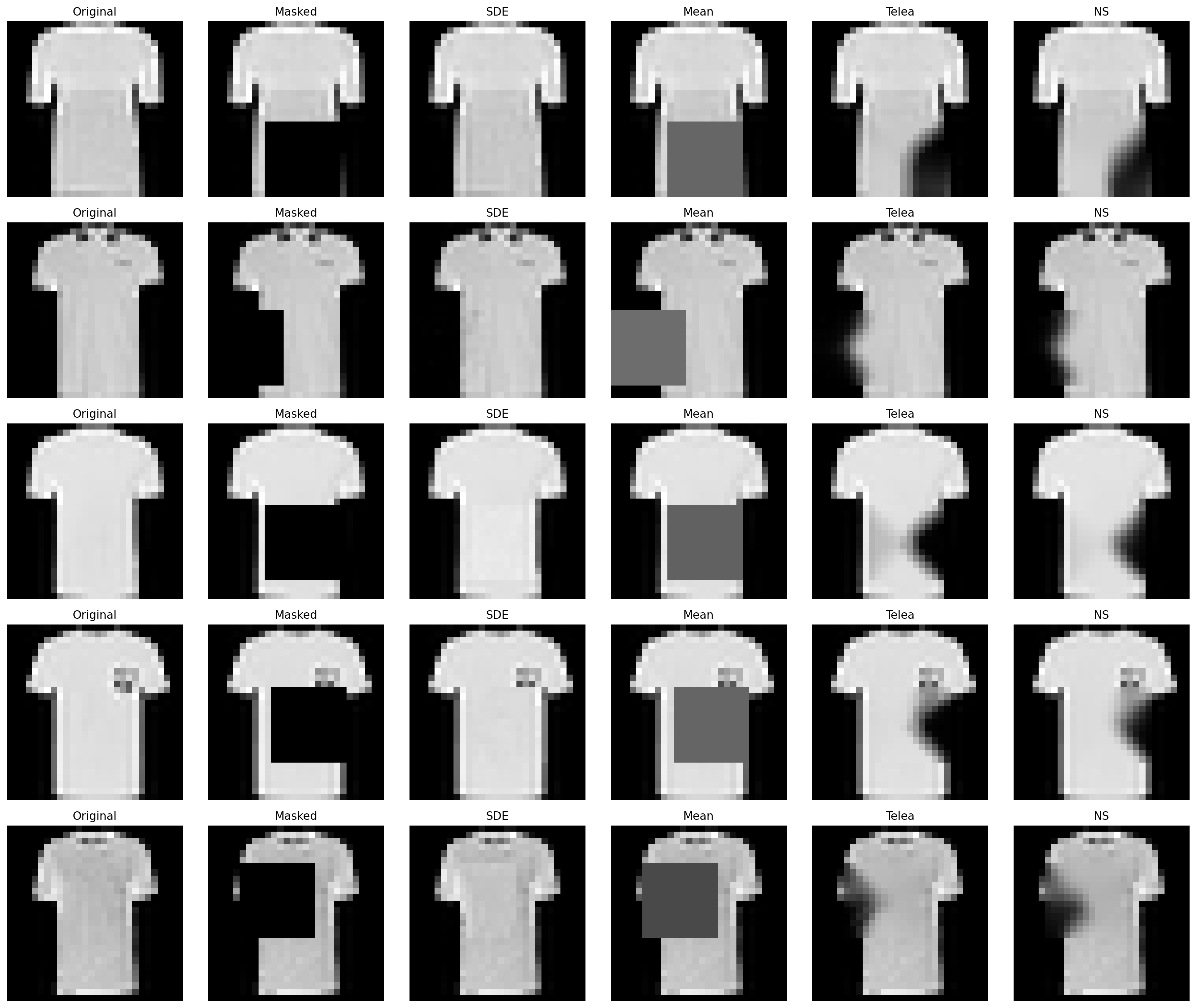}
    \caption{Label "T-shirt/top" Comparison of Image Inpainting Performance of SDE, Mean-Fill, Telea, and NS in Fashion-MNIST}
    \label{fashion_0}
\end{figure}

\begin{figure}[H]
    \centering
    \includegraphics[width=1\textwidth]{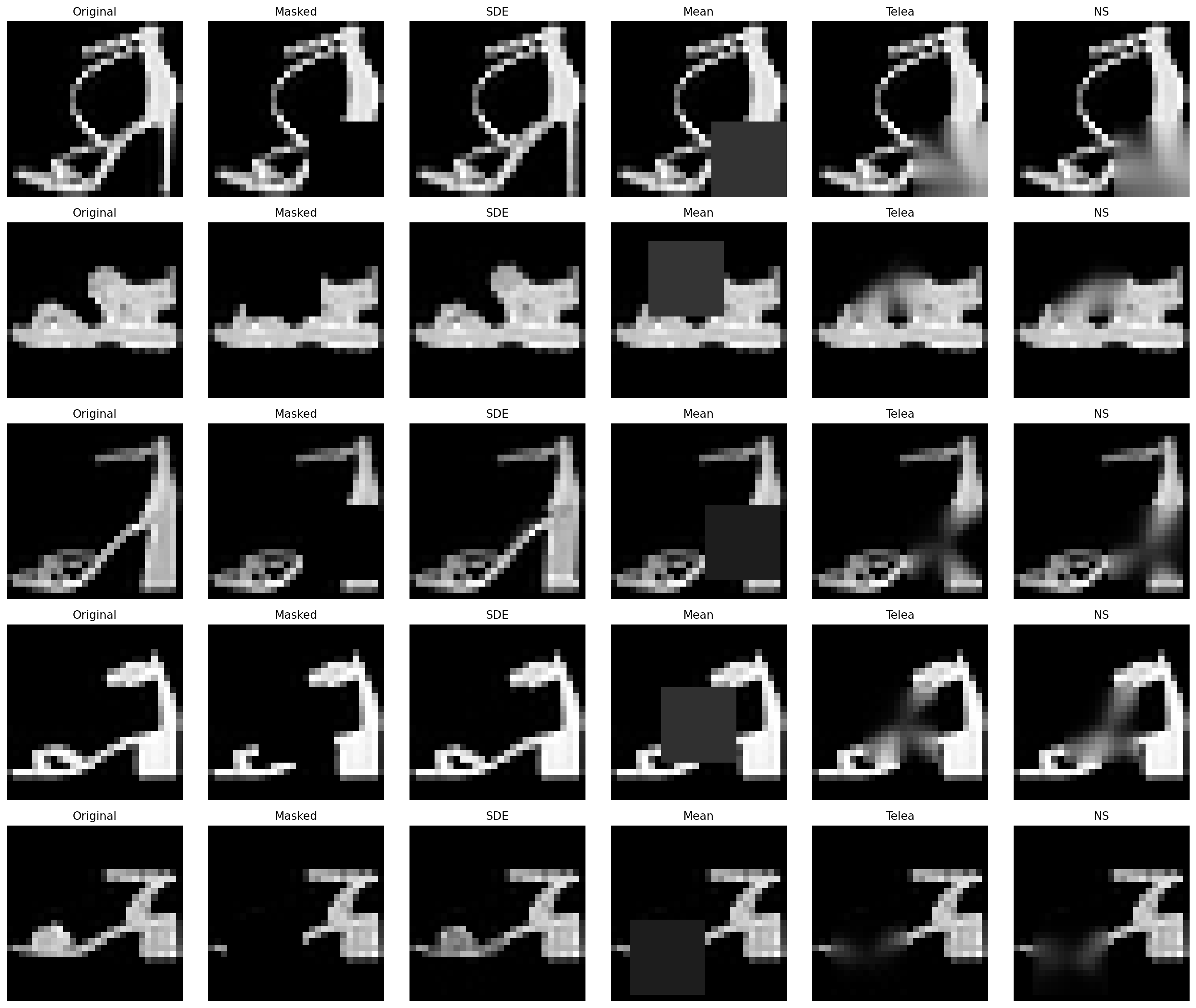}
    \caption{Label "Sandal" Comparison of Image Inpainting Performance of SDE, Mean-Fill, Telea, and NS in Fashion-MNIST}
    \label{fashion_5}
\end{figure}

\begin{figure}[H]
    \centering
    \includegraphics[width=1\textwidth]{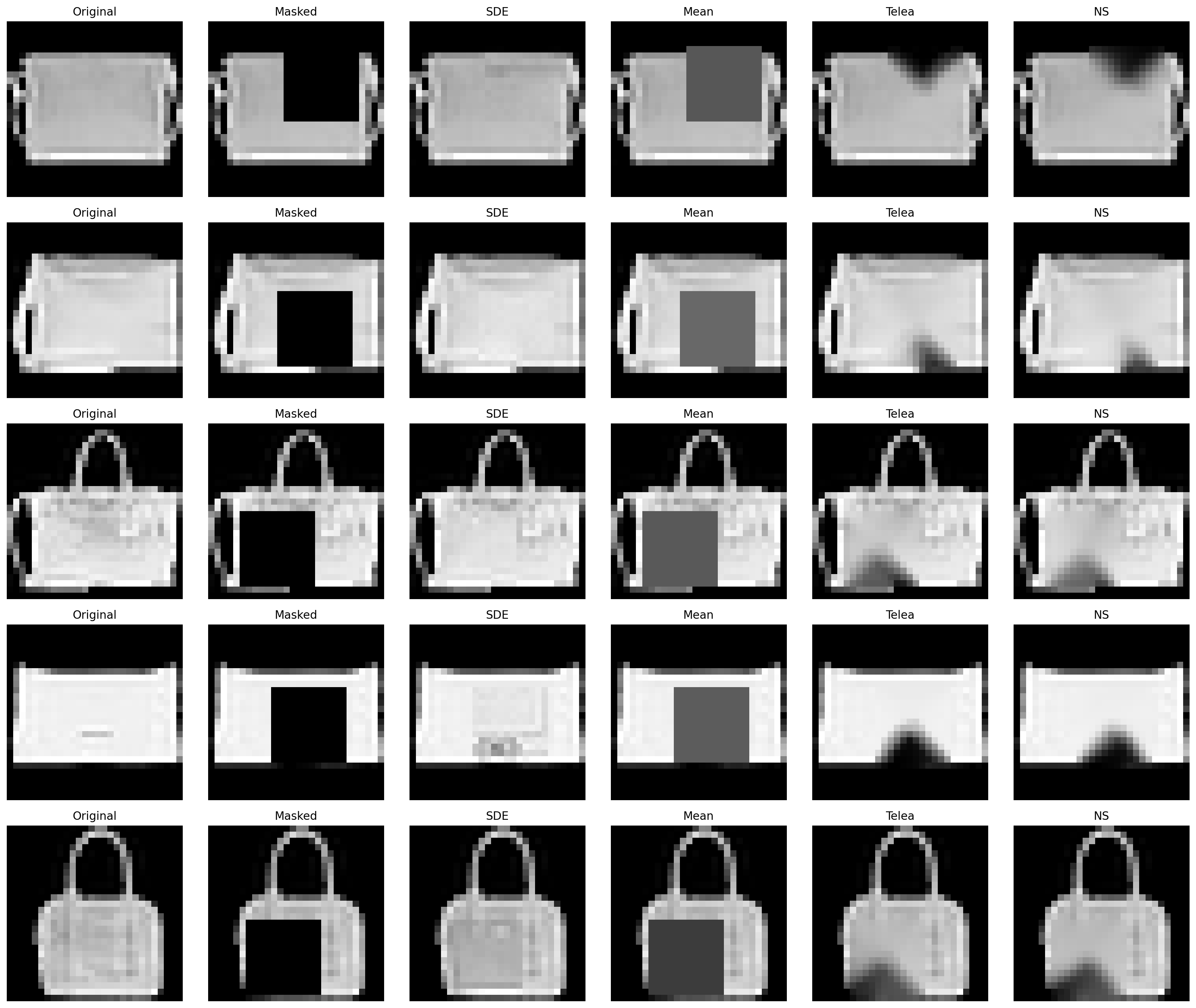}
    \caption{Label "Bag" Comparison of Image Inpainting Performance of SDE, Mean-Fill, Telea, and NS in Fashion-MNIST}
    \label{fashion_8}
\end{figure}

\end{document}